%% file: main.tex
\documentclass[11pt]{article}
\usepackage[linesnumbered,ruled,vlined]{algorithm2e}
\usepackage{amsfonts}
\usepackage{amsthm}
\usepackage{amsmath,amssymb}
\usepackage{arydshln}
\usepackage{authblk}
\usepackage{bbm}
\usepackage{bm}
\usepackage{booktabs}
\usepackage{enumitem}
\usepackage[margin=1in]{geometry}
\usepackage{graphicx}
\usepackage{hhline}
\usepackage{mathtools}
\usepackage{multirow}
\usepackage{xcolor}
\usepackage[colorlinks=true,citecolor=blue]{hyperref}

\newtheorem{theorem}{Theorem}[section]

\newtheorem{assumption}{Assumption}

\graphicspath{{figs/}}

\title{Federated scientific machine learning for approximating functions and solving differential equations with data heterogeneity}
\author[1,\dag]{Handi Zhang}
\author[2,\dag]{Langchen Liu}
\author[2,3,*]{Lu Lu}
\affil[1]{Graduate Group in Applied Mathematics and Computational Science, University of Pennsylvania, Philadelphia, PA 19104, USA}
\affil[2]{Department of Statistics and Data Science, Yale University, New Haven, CT 06511, USA}
\affil[3]{Wu Tsai Institute, Yale University, New Haven, CT 06510, USA}
\affil[$\dag$]{These authors contributed equally to this work.}
\affil[*]{Corresponding author. Email: lu.lu@yale.edu}
\date{}

\begin{document}
\maketitle

\begin{abstract}
By leveraging neural networks, the emerging field of scientific machine learning (SciML) offers novel approaches to address complex problems governed by partial differential equations (PDEs). In practical applications, challenges arise due to the distributed essence of data, concerns about data privacy, or the impracticality of transferring large volumes of data. Federated learning (FL), a decentralized framework that enables the collaborative training of a global model while preserving data privacy, offers a solution to the challenges posed by isolated data pools and sensitive data issues. Here, this paper explores the integration of FL and SciML to approximate complex functions and solve differential equations. We propose two novel models: federated physics-informed neural networks (FedPINN) and federated deep operator networks (FedDeepONet). We further introduce various data generation methods to control the degree of non-independent and identically distributed (non-iid) data and utilize the 1-Wasserstein distance to quantify data heterogeneity in function approximation and PDE learning. We systematically investigate the relationship between data heterogeneity and federated model performance. Additionally, we propose a measure of weight divergence and develop a theoretical framework to establish growth bounds for weight divergence in federated learning compared to traditional centralized learning. To demonstrate the effectiveness of our methods, we conducted 10 experiments, including 2 on function approximation, 5 PDE problems on FedPINN, and 3 PDE problems on FedDeepONet. These experiments demonstrate that proposed federated methods surpass the models trained only using local data and achieve competitive accuracy of centralized models trained using all data.
\end{abstract}


\input{content/Introduction}
\input{content/Methods}
\input{content/Theory}
\input{content/FunctionApprox}
\input{content/PINN}
\input{content/OperatorLearning}
\input{content/Conclusion}

\section*{Acknowledgments}
This work was supported by the U.S. Department of Energy Office of Advanced Scientific Computing Research under Grants No.~DE-SC0025592 and No.~DE-SC0025593, and the U.S. National Science Foundation under Grant No.~DMS-2347833.

\appendix
\input{content/Appendix}

\bibliographystyle{unsrt} 
\bibliography{main}

\end{document}

%% file: content/Introduction.tex
\section{Introduction}

With the development of big data and the growth of computing resources, machine learning has been widely used to perform classification, clustering, and regression tasks, and tackle complex problems across scientific fields. The universal approximation theorem of neural networks has established a solid theoretical basis for deep learning. This has led to the widespread adoption of neural networks for solving partial differential equations (PDEs) in scientific machine learning (SciML), in contrast to traditional numerical methods such as the finite difference and finite element methods, which are computationally expensive. 

Physics-informed neural networks (PINNs)~\cite{raissi2019physics, lu2021deepxde, karniadakis2021physics} have emerged as a groundbreaking framework within SciML, offering a novel solution to both forward and inverse problems governed by PDEs. By integrating the PDE loss into the neural network's loss function, PINNs enable effective and accurate handling of these complex problems. Despite the great potential, PINNs still require further enhancements to improve prediction accuracy, computational efficiency, and training robustness. To address some of the challenges in practice, researchers have proposed a series of extensions to the vanilla PINN framework, targeting performance improvement from various angles. For instance, the use of meta-learning has led to the discovery of better loss functions~\cite{psaros2022meta}, and the development of gradient-enhanced PINNs has enabled the incorporation of gradient information from the PDE residual into the loss function~\cite{yu2022gradient}. Different methods have been devised to automatically adjust the weights of these loss terms to optimize the performance, effectively balancing the contributions of PDE and initial/boundary conditions~\cite{mcclenny2020self, jin2021nsfnets}. Other techniques, such as residual-based adaptive sampling~\cite{wu2023comprehensive} and multiscale Fourier features~\cite{wang2021eigenvector} were also applied to improve PINN accuracy. User-friendly libraries and toolboxes have also been developed to facilitate further research in PINNs~\cite{lu2021deepxde,hao2023pinnacle}.

In addition to approximating functions, neural networks can also approximate the mappings between infinite-dimensional function spaces, i.e., operator learning. The deep operator network (DeepONet) has exhibited impressive performance in effectively approximating the operators involved in various PDE~\cite{lu2019deeponet, deeponetNatureML}. Once the network is trained, it only requires a forward pass to derive the PDE solution for a new input, even in the extrapolation scenario~\cite{zhu2023reliable}. DeepONet has found successful applications in diverse fields such as multiscale bubble dynamics~\cite{lin2021operator}, brittle fracture analysis~\cite{GOSWAMI2022114587}, solar-thermal systems forecasting~\cite{osti_1839596}, electroconvection~\cite{cai2021deepm}, and fast multiscale modeling~\cite{yin2022interfacing}. Moreover, researchers have proposed several extensions of DeepONet in recent studies, including DeepONet incorporating proper orthogonal decomposition~\cite{lu2022comprehensive}, physics-informed DeepONet~\cite{pideeponet}, multifidelity DeepONet~\cite{lu2022multifidelity}, multiple-input DeepONet~\cite{jin2022mionet}, Fourier-DeepONet~\cite{zhu2023fourier, jiang2024fourier}, and DeepONet integrated with uncertainty quantification~\cite{uqdeeponet}. Recently, the distributed approaches of DeepONet have also been studied by researchers, such as D2NO~\cite{zhang2023d2no} to handle heterogeneous input function spaces.

However, although SciML has significant achievements and diverse applications across various domains, challenges arise in real-world applications due to the distributed essence of data and privacy issues. A relevant example is atmospheric science, where extensive data from observations and numerical models is used. People obtain large volumes of observations from remote sensing (e.g., satellites and radars) and conventional platforms (e.g., weather stations, aircraft, radiosondes, ships, and buoys) from different devices and store the data in various forms at different agencies. Unfortunately, most existing SciML models do not apply to these critical situations due to the absence of capabilities in decentralized learning. In some other cases, the underlying data could be privacy-sensitive and thus might be restricted from being aggregated to protect proprietary information. This concern is also common in subsurface energy applications. Proprietary technology, confidentiality clauses, and regulatory requirements typically limit data sharing while concerns surrounding reserve estimates, geopolitical sensitivity, and operational risks reinforce the need for data privacy. To address these issues, researchers explore the distributed training methods that utilize data across different resources while maintaining privacy for private domain data. With this motivation, federated learning (FL), introduced by McMahan et al.~\cite{mcmahan2017communication}, comes to play an essential role. 

Federated learning is a decentralized ML setting, where many clients collaboratively train a common ML model under the orchestration of a central server while keeping the training data decentralized~\cite{mcmahan2017communication}. In this framework, while preserving data privacy by keeping the data on clients, significant reductions in data transfer volume are also achieved by coordinating updates with a central server via aggregation and broadcasting, leading to lower communication costs and better robustness. Federated learning has been widely used in privacy-sensitive fields, such as healthcare~\cite{lee2018privacy, brisimi2018federated, li2019privacy, huang2019patient}, finance~\cite{long2020federated,byrd2020differentially, toosi2012financial} and Internet of things~\cite{nguyen2021federated}. This essential direction emphasizes how to prevent indirect leakage of local data and thus preserve privacy. For example, differential privacy is widely used for data privacy protection~\cite{wei2020federated, truex2020ldp}. Geyer et al.\ introduced randomized mechanisms such as random sub-sampling and distorting to protect client's data~\cite{geyer2017differentially}. In real applications of federated learning, another critical aspect is communication efficiency, which directly inspires several communication efficient algorithms including Federated Averaging (FedAvg)~\cite{mcmahan2017communication}, FedAvg with momentum~\cite{hsu2019measuring}, and adaptive federated optimizers (FedAdagrad, FedAdam, FedYogi)~\cite{reddi2020adaptive}. 

This paper explored the novel applications of federated learning in SciML, including function approximation, solving PDEs, and learning operators. Specifically, we proposed the federated averaging algorithm with local Adam update (FedAvg-Adam) and formulated the federated versions of PINN (FedPINN) and DeepONet (FedDeepONet) to solve scientific problems. Moreover, we proposed to utilize 1-Wasserstein distance ($W_1$) to quantify the level of non-independent and identically distributed (non-iid) nature of client data in approximation tasks and solving PDEs with PINNs. To simulate the non-iid setting in regression tasks, we proposed various data generation methods for one- and two-dimensional problems, which partition the original dataset into a certain number of subsets and then distribute to two or multiple clients for federated training. Unlike the domain partition method for generating non-iid datasets for function approximation and FedPINN, we control the non-iid level in operator learning by creating different functional spaces to sample the functions. Then, to quantitatively study the differences between federated learning and traditional centralized learning, we introduce the weight divergence to measure the difference between models and theoretically analyze the upper bound of the weight divergence under several assumptions.

In the numerical experiments, we demonstrate that the federated model outperforms the extrapolating models and gradually approaches the centralized model with a smaller $W_1$ distance and less data heterogeneity between clients. This phenomenon not only applies to the function approximation tasks and PINNs but also generalizes to operator learning. Besides, empirical results show that the weight divergence positively correlates with the non-iid level and thus explains the performance degradation of federated models when the data are highly heterogeneous. For operator learning, another interesting result is that the FedDeepONet is insensitive to communication efficiency.

The paper is organized as follows. In Section~\ref{sec:method}, after briefly introducing the fundamental idea of federated scientific machine learning (FedSciML), we present the FedAvg-Adam used in the training procedure, different data generation methods, and the 1-Wasserstein distance as the measurement of data heterogeneity. Subsequently, in Section~\ref{sec:theory}, we introduce the definition of weight divergence and then provide a theoretical analysis of the growth bound of weight divergence between the FedSciML model and the centralized model. In Section~\ref{sec:func_approx}, we provide experiments for function approximation tasks, including 1D/2D functions with two clients and 2D functions learned by multi-clients. In Section~\ref{sec:pinn}, we test the performance of proposed methods for various types of PDEs that cover both forward and inverse problems, including the Poisson equation, Helmholtz equation, Allen-Cahn equation, the inverse problem of 2D Navier-Stokes equation and the inverse problem of inferring the space-dependent reaction rate in a diffusion-reaction system. In Section~\ref{sec:deeponet}, after introducing the architecture of DeepONet, we test FedDeepONet on learning the antiderivative operator, diffusion-reaction equation, and burgers equation. Moreover, we explore DeepONet's insensitivity property concerning different numbers of local iterations. Finally, we conclude the paper and discuss some future directions in
Section~\ref{sec:conclusion}.

%% file: content/Methods.tex
\section{Methods}
\label{sec:method}

This section briefly explains federated scientific machine learning, including the non-iid setting of federated learning and the federated averaging algorithm for training (Sec.~\ref{subsec:fedsciml}). Additionally, we introduce the concept of 1-Wasserstein distance for quantifying data heterogeneity in regression tasks, and we provide different methods for partitioning and assigning data to simulate the non-iid distribution for local datasets (Sec.~\ref{subsec:date_gen}).

\subsection{Federated scientific machine learning}
\label{subsec:fedsciml}

\begin{figure}[htbp]
    \centering
    \includegraphics[width=1.0\textwidth]{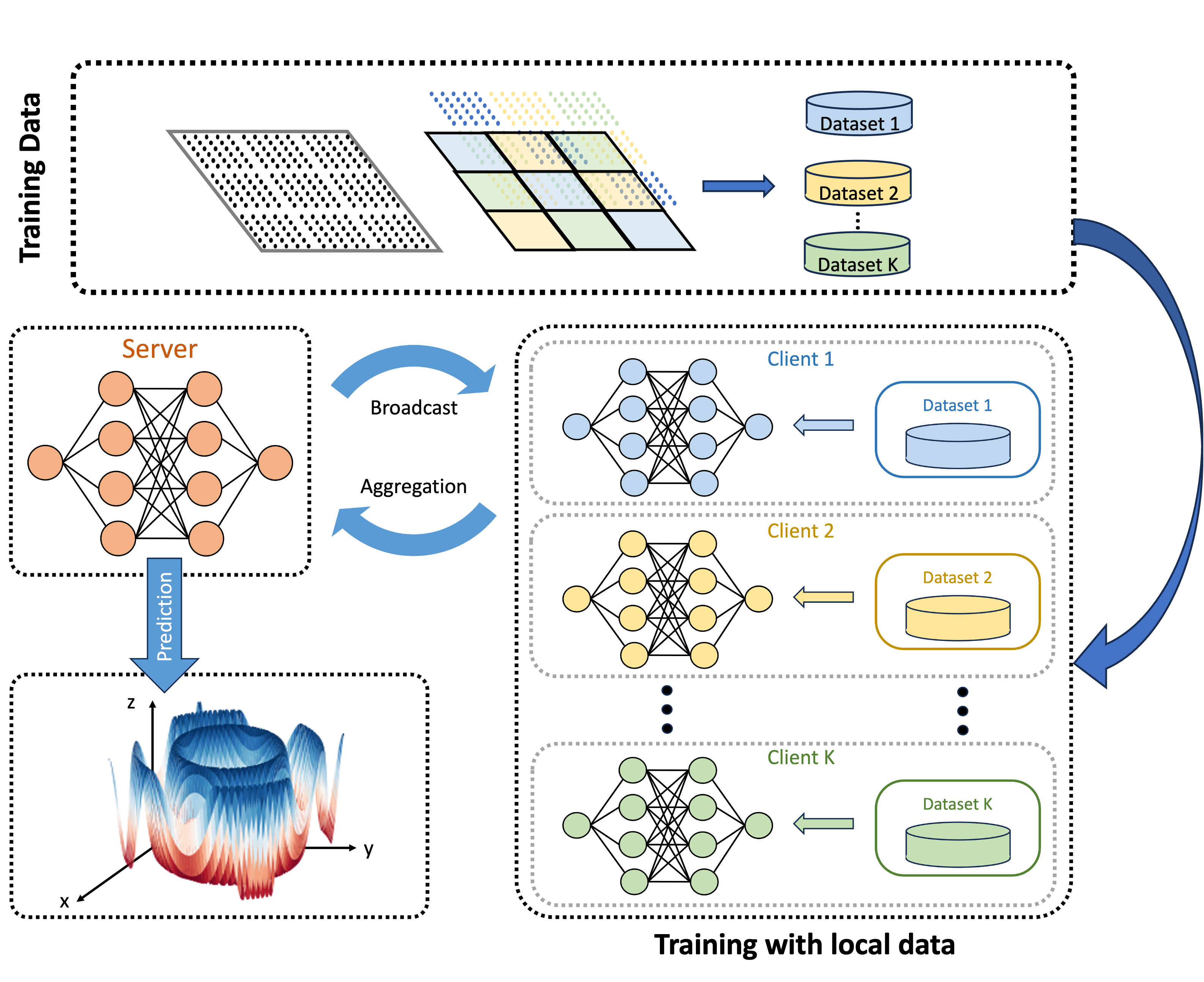}
    \caption{\textbf{Workflow of federated scientific machine learning for approximating operators and solving differential equations.} (\textbf{Top}) In FedSciML, the training data is distributed across $K$ clients. (\textbf{Bottom}) Each client has its model and dataset. The models are trained through a collaborative training procedure, which includes (1) the aggregation from local models to the server model and (2) the broadcast from the server model back to local models.}
    \label{fig:workflow}
\end{figure}

Federated learning is a method that allows multiple devices or clients to collaboratively learn a global model while keeping their raw data localized. In this framework, each participating client has its private dataset and updates the global model by performing local computations on the data and exchanging model updates with a central server. An application of this concept is federated scientific machine learning (FedSciML), which applies federated learning to scientific machine learning problems.

We illustrate the workflow of FedSciML in Fig.~\ref{fig:workflow}. FedSciML aims to learn a function, a PDE, or an operator (probably with data heterogeneity) by training on multiple local datasets stored in local clients without direct communications. In this study, we focus on data heterogeneity. First, a training dataset is generated using proposed methods and distributed among numerous clients to simulate the heterogeneous and decentralized setting. Then, each client interacts with other clients indirectly via a collaborative training procedure. This procedure consists of an aggregation stage from local clients to the server and a subsequent broadcast stage from the server back to local clients. Generally, federated learning algorithms exchange the parameters (i.e., the weights and biases of the neural networks) between models at specific communication frequencies to train the server model. Because of the implicit usage of local datasets, the server model demonstrates satisfactory generalizability while preserving local data privacy.

The following sections briefly overview the learning objectives in federated scientific machine learning and introduce different types of non-iid data in applications. We then present the federated averaging algorithm with local Adam update (FedAvg-Adam) for collaboratively training a global model with data distributed across multiple local clients. It is important to note that Fed-DeepONet~\cite{Moya_2022} already combines FL and DeepONet. Our work extends the incorporation of FL to PINNs and function approximation problems, further examining the impact of data heterogeneity on FedSciML.

\subsubsection{Problem formulation}

In federated learning, we formulate the optimization problem as
$$\min_{\theta} \mathcal{L}(\theta)=\sum_{k=1}^K\frac{N_k}{N}\mathcal{L}_k(\theta)\qquad\text{with} \qquad \mathcal{L}_k(\theta)=\frac{1}{N_k}\sum_{d_i\in{D_k}}\ell(\theta; d_i),$$
where $\mathcal{L}$ and $\mathcal{L}_k$ are the federated server loss and the $k$-th client loss. $K$ denotes the number of clients. $D_k$ denotes the dataset of client $k$ and $N_k=|D_k|$ is the number of data of client $k$. $N =\sum_k N_k$ is the total number of data, and $\ell(\theta; d_i)$ is the pointwise loss of the prediction based on data sample $d_i$ made by weights $\theta$. Specifically, in FedSciML, we usually take $\ell $ as the squared error loss so that $\mathcal{L}_k$ measures the mean square error of the client $k$. 

The key idea of federated learning is to accommodate the challenges arising from non-iid data distributions across clients, communication constraints, and heterogeneous computing resources. Previous studies have examined the classification of non-iid data systems, encompassing diverse client distributions and violations of independence, and various algorithms have been proposed to address non-iid data (see a review paper~\cite{kairouz2021advances}). Let $P_i(x)$ and $P_i(y)$ denote the probability distribution of the feature and label data for client $i$, respectively, and $P_i(\cdot|\cdot)$ represent the conditional probability. We illustrate four types of non-iid data caused by non-identical distributions in Table~\ref{tab:noniid}. It is worth noting that, in addition to heterogeneous probability distributions, both quantity skew and dataset imbalance can also result in non-identical data. This work primarily focuses on data heterogeneity caused by covariance shifts.

\begin{table}[htbp]
\centering
\caption{\textbf{Type of non-IID data.}}
\begin{tabular}{c|cc}
\toprule
Type & Distribution\\
\midrule
Covariance shift & $P_i(x)\ne P_j(x),\ P_i(y|x)=P_j(y|x)$\\
Prior probability shift &$P_i(y)\ne P_j(y),\ P_i(x|y)=P_j(x|y)$ \\
Concept drift  &$ P_i(y)=P_j(y),\ P_i(x|y)\ne P_j(x|y)$\\
Concept shift  &$P_i(x)=P_j(x),\ P_i(y|x)\ne P_j(y|x)$\\
\bottomrule
\end{tabular}
\label{tab:noniid}
\end{table}

\subsubsection{Federated averaging algorithm with local Adam update (FedAvg-Adam)}

Let us consider a random fraction of clients, denoted as $C$, to receive the current model parameters. Additionally, $\eta$ represents a fixed learning rate for each client $k$, and $E$ is the number of local epochs for each client. In our numerical experiments, we set $C$ to 100\% to ensure the participation of all clients in the communication round, meaning there will be one broadcast and one aggregation. 

The FedAvg algorithm~\cite{mcmahan2017communication} consists of mainly two stages. At the $l$-th global epoch, the algorithm performs the following actions:
\begin{enumerate}[label=(\alph*)]
    \item Server-to-client broadcasting of the current server model: \begin{equation*}
        \theta_k^{l,0} = \theta^{l},\quad\text{for }k=1,\cdots, K.
    \end{equation*}
    \item Local gradient computation for $E$ steps: \begin{equation*}
        g_k^{i}=\nabla \mathcal{L}_k(\theta^{l,i-1}_k),\quad\text{for }k=1,\cdots, K\text{ and } i=1,\cdots, E,
    \end{equation*}
    \item Global aggregation of all locally updated results:
    \begin{equation} \label{eq:aggre}
       \theta^{l+1}\leftarrow \theta^l-\eta\sum_{k=1}^K\sum_{i=1}^E\frac{N_k}{N}g_k^{i}.
    \end{equation}
\end{enumerate}
Here, $\theta^{l, i}_k$ denotes the local model for client $k$ at global epoch $l$ and local epoch $i$, and $\theta^l$ denotes the global or server model at global epoch $l$. 

In the generalized FedAvg method (also known as FedOPT)~\cite{reddi2020adaptive}, the federated update procedure extends to various methods to perform local update and aggregation, where the extension consists of Client-OPT and Server-OPT. Each local client will perform the local update by
\begin{equation*}
    \theta_k^{l,i+1}=\text{Client-OPT($\theta_k^{l,i}, g_k^{i+1}, \eta, l$)},\quad\text{for }i=0,\cdots, E-1.
\end{equation*}
This gives the total local change by
\begin{equation*}
    \Delta_k^l =\theta_k^{l,E} - \theta_k^{l,0}.
\end{equation*}
The aggregation step is then defined using the difference in local model parameters
\begin{equation}\label{eq:fedopt_aggre}
    \Delta^{l+1}=\sum_{k=1}^K\frac{N_k}{N}\Delta_k^{l},\qquad \text{and}\qquad\theta^{l+1}=\text{Server-OPT($\theta^l, -\Delta^{l+1},\eta,l$)}.
\end{equation}

In the context of this general framework, vanilla FedAvg (Eq.~\eqref{eq:aggre}) is a special case of Fed-OPT where both the Client-OPT and Server-OPT (Eq.~\eqref{eq:fedopt_aggre}) implement gradient descent. The empirical results presented in this paper show that the Adam optimizer performs better than SGD. In our study, we introduce FedAvg-Adam (Algorithm~\ref{algo:fedavg}), in which we replace gradient descent with the Adam optimizer for updating the local model in all experiments. We then average the change of all local clients to update the global model. Additionally, it is easy to show that our algorithm is equivalent to Fed-OPT, with Adam as the Client-OPT optimizer and gradient descent as the Server-OPT optimizer. In practice, we averaged the local model directly instead of averaging the change to update the server model.

\begin{algorithm}[htbp]
\label{algo:fedavg}
\DontPrintSemicolon
\SetAlgoLined
\SetKwInput{KwInput}{Input}
\SetKwInput{KwOutput}{Output}
\SetKwFor{For}{for}{do}{end}
\SetKwIF{If}{ElseIf}{Else}{if}{then}{else if}{else}{endif}

\KwInput{Initial model $\theta^0$; $K$ clients indexed by $k$; local epochs $E$; learning rate $\eta$}
\For{$l=0,1,2\cdots $}{
    \For{$k=1,\cdots, K $ \textbf{in parallel}}{
        Initialize local model $\theta_k^{l,0} \leftarrow \theta^{l}$\;
        \For{$i=1,\cdots,E $}{
            Compute the local stochastic gradient $g_k(\theta_k^{l,i-1})$\;
            $\theta_k^{l,i} \leftarrow$ Adam$(g_k^i, \eta)$\;
        }
        Compute the total local change $\Delta_k^l \leftarrow \theta_k^{l,E} - \theta_k^{l,0}$\;
    }
    Do global aggregation $\theta^{l+1} \leftarrow \theta^l + \sum_{k=1}^K \frac{N_k}{N}\Delta_k^l$, or equivalently $\theta^{l+1} \leftarrow \sum_{k=1}^K \frac{N_k}{N}\theta_k^{l, E}$\;
}
\caption{FedAvg-Adam.}
\end{algorithm}




\subsection{Data generation and quantification of data heterogeneity}
\label{subsec:date_gen}
Before detailing our proposed data generation methods, we introduce the Wasserstein distance as a metric for assessing data heterogeneity in regression tasks. To simulate a federated setting for function approximation, solving PDEs, or learning operators, we develop corresponding data generation methods for one-dimensional and two-dimensional cases, addressing scenarios with two and multiple clients. For function approximation and differential equation solving, we partition the original dataset into several subdomains along the axes and distribute them to local clients (Sec.~\ref{subsubsec:1d} and Sec.~\ref{subsubsec:2d}). Additionally, in the context of operator learning—where the model is trained on multiple functions sampled from a specific functional space—we introduce an alternative approach to quantify and manage the level of non-iid by generating functions from different functional spaces (Sec.~\ref{subsubsec:operator}).
\subsubsection{1-Wasserstein distance for quantifying data heterogeneity}

To quantify the difference between the distributions of clients, we consider the 1-Wasserstein distance. Let $(M,d)$ denote a metric space. The general form of the Wasserstein distance between two probability measures $\mu$ and $\nu$ on $M$ with finite $p$-moments is given by
\begin{equation*}
    W_p(\mu, \nu) = \left(\inf_{\gamma \in \Gamma(\mu, \nu)} \int d(x,y)^pd\gamma(x,y)\right)^{1/p},
\end{equation*}
where $\Gamma(\mu,\nu)$ is the set of all couplings of $\mu$ and $\nu$. We can simplify the 1-Wasserstein distance when $p = 1$ to 
\begin{equation}
\label{eq:w1}
W_1(\mu, \nu) = \inf_{\gamma \in \Gamma(\mu, \nu)} |x-y|d\gamma(x,y),
\end{equation}
where the coupling $\gamma$ is the joint probability measure on $M\times M$ with marginals $\int_M\gamma(x,y)dy=\mu(x)$, $\int_M\gamma(x,y)dx=\mu(y)$.  

For $p=1$, $W_1$ is equivalent to the Earth Mover's Distance (EMD). In the subsequent implementation, rather than directly working with Eq.~\eqref{eq:w1}, we compute the discretized $W_1$ distance to measure the level of non-iid in datasets by using the Python Optimal Transport package~\cite{flamary2021pot}. Specifically, consider discrete samples obtained through random sampling or uniform grids from $\mu$ and $\nu$ with finite dimensions. We have two sample distributions $\mu_{n_1}$ and $\nu_{n_2}$, where $n_1$ and $n_2$ represent the number of sample points in each distribution. The POT package then computes the $W_1$ distance by solving the EMD problem, defined as
\begin{align*}\begin{aligned}W_1(\mu_{n_1},\nu_{n_2})=\min_\gamma \quad \langle \gamma, \mathbf{M} \rangle_F,\\s.t. \ \gamma \mathbf{1} = \mathbf{a},\ \gamma^T \mathbf{1} = \mathbf{b},\ \gamma \geq 0,\end{aligned}\end{align*}
where $\mathbf{M} \in \mathbb{R}^{n_1 \times n_2}$ is the Euclidean distance matrix between $\mu_{n_1}$ and $\nu_{n_2}$, and $m_{ij}$ represents the distance between the $i$-th data point from the first distribution and the $j$-th data point from the second distribution. $\mathbf{a} \in \mathbb{R}^{n_1}$ and $\mathbf{b} \in \mathbb{R}^{n_2}$ are the sample weights of $\mu_{n_1}$ and $\nu_
{n_2}$. The optimization problem seeks to solve the minimal cost of transport between the two distributions, subject to the cost matrix $\mathbf{M}$ and the sample weights $\mathbf{a}$ and $\mathbf{b}$.

When there are three or more clients, i.e., $K \geq 3$, we compute the mean pairwise $W_1$ distance to quantify the data heterogeneity among all clients as
\begin{equation}
\label{eq:pairw1}
    \overline{W_1} = \frac{1}{(K-2)(K-1)}\sum_{i=1}^{K-1} \sum_{j=i+1}^{K} W_1(\mu_i, \mu_j).
\end{equation}
\subsubsection{1D data generation}
\label{subsubsec:1d}

In the context of function approximation and solving differential equations, suppose the total number of data points is $N$. We aim to partition the original dataset into smaller subdomains and assign them to $K$ local clients.

\begin{figure}[htbp]
    \centering
    \includegraphics[width=0.9\textwidth]{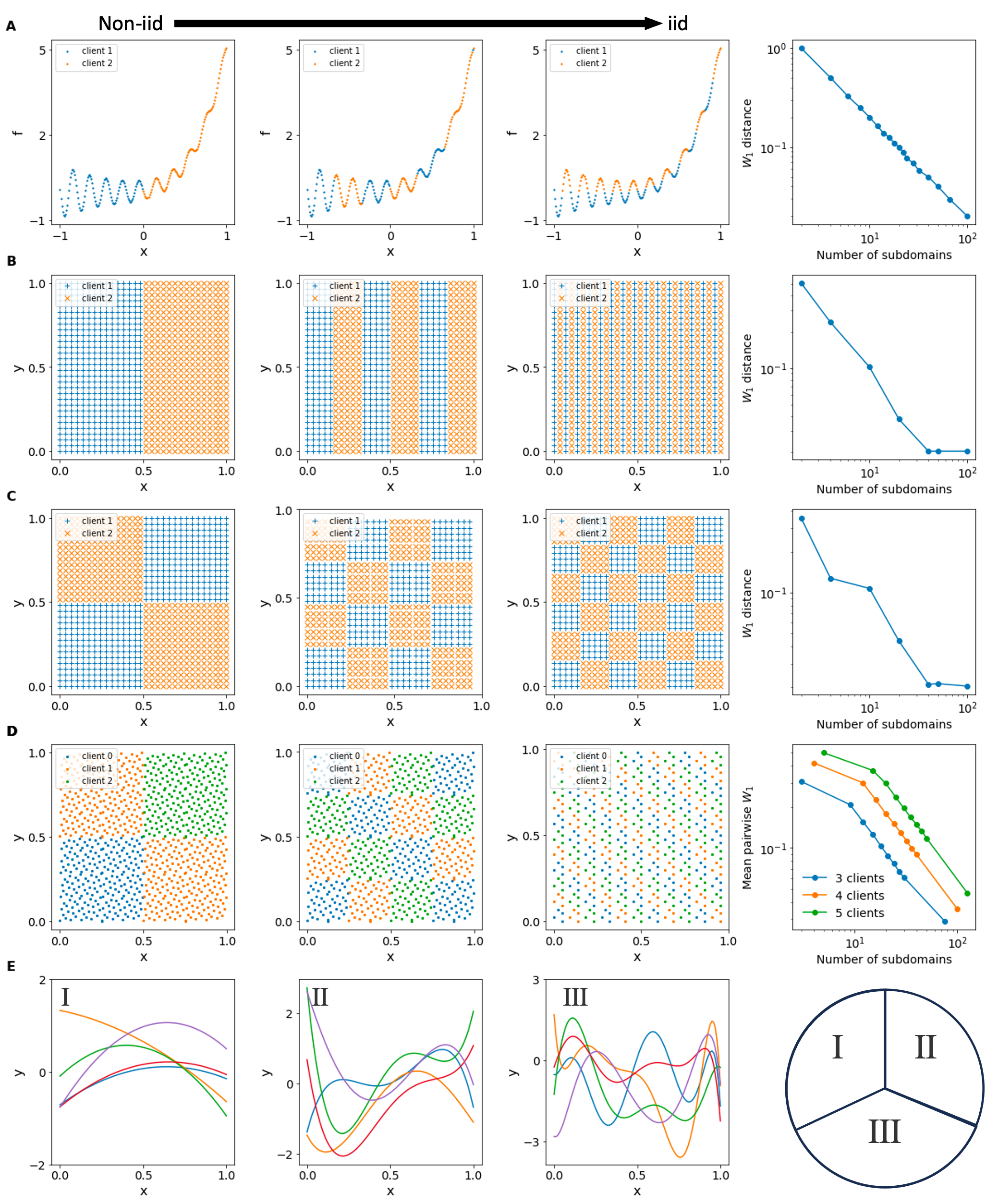}
    \caption{\textbf{Visualization of data generation methods.} (\textbf{A}) Data generation for 1D data with two clients in different data heterogeneity. From left to right, the level of iid increases. The last column is the change of $W_1$ distance for different numbers of subdomains. (\textbf{B}) 2D data with $x$-partition and two clients. (\textbf{C}) 2D data with $xy$-partition and two clients. (\textbf{D}) 2D data with $xy$-partition and three clients. The last column shows the mean pairwise $W_1$ distance change for three, four, and five clients. (\textbf{E}) Visualization of a data generation method for operator learning. From the first column to the third column, functions are sampled from different functional spaces.}
    \label{fig:assign}
\end{figure}

In the 1D case, let $\{x_i \in \mathcal{I},\, i = 1, \dots, N\}$ represent $N$ data points sampled from an interval $\mathcal{I} \subset \mathbb{R}$. Let $n$ denote the number of subdomains assigned to each client. The dataset is first divided into $n$ subdomains, after which each client alternately receives data, as illustrated in the first column of Fig.~\ref{fig:assign}A. The resulting local datasets are denoted as $\{\Omega_k\}_{k=1}^K$, with $|D_k| = N_k$. If $N \mod (nK) \ne 0$, the remaining sample points are sequentially assigned to the local clients. The last column in Fig.~\ref{fig:assign}A shows the variation in the $W_1$ distance, calculated from Eq.~\eqref{eq:w1}, as the number of subdomains $n$ increases. This demonstrates that increasing the number of partitions reduces the $W_1$ distance in scenarios with two clients, indicating a higher degree of iid.

\subsubsection{2D data generation}
\label{subsubsec:2d}

For the 2D dataset, we adopt two distinct methods for assigning data to clients: the $x$-partition and the $xy$-partition. Given $N$ data points sampled from a 2D region $\mathcal{D} \subset \mathbb{R}^2$, denoted as $\{(x_i, y_i) \in \mathcal{D},\, i=1,\dots, N\}$, along with the number of clients $K$ and the number of subdomains $n$ per axis, we define the data generation methods as follows.
\begin{itemize}
    \item The $x$-partition method divides the $x$ coordinate into $n$ subintervals while leaving the $y$ coordinate unchanged. This approach is analogous to the 1D data generation process, as it operates exclusively on the first axis (Fig.~\ref{fig:assign}B).
    
    \item The $xy$-partition method divides both the $x$ and $y$ coordinates into $n$ subintervals, resulting in $n^2$ subdomains. These subdomains are then assigned to clients in an alternating manner. Fig.~\ref{fig:assign}C illustrates the $xy$-partition data generation for two clients, while Fig.~\ref{fig:assign}D depicts the case involving multiple clients.
\end{itemize}

We compare the disparities between the two methods of generating 2D data and detail the corresponding changes in $W_1$ in the final column of Figs.~\ref{fig:assign}B--D. For the $xy$-partition generation method, we generate all datasets from the entire domain using Hammersley sampling, a low-discrepancy method based on the Hammersley sequence. Our results indicate that the difference in $W_1$ between the $xy$-partition and $x$-partition methods becomes negligible as the number of subdomains increases. Therefore, the data generation approach in subsequent experiments may depend on the specific problem setting. For the 2D time-independent problem, we employ the $xy$-partition method. Conversely, for the time-dependent problem with 1D in space, the $x$-partition method in space coordinate is more appropriate, as it allows each client to be distinguished based solely on the spatial domain.

Our proposed data generation method is also adaptable to multiclient experiments. In the multiclient scenario, we employ the mean pairwise $W_1$ distance, as defined in Eq.~\eqref{eq:pairw1}, as a measure of data heterogeneity. We visualize the data distributions and compute the mean pairwise $W_1$ for three, four, and five clients (Fig.~\ref{fig:assign}D), demonstrating that the multiclient case follows a similar trend to the two-client case. However, the non-iidness becomes more pronounced as the number of clients increases.

\subsubsection{Data generation for operator learning}
\label{subsubsec:operator}

Operator learning aims to learn linear and nonlinear operators that map input functions to output functions, i.e., the mapping between two infinite-dimensional functional spaces. In the federated operator learning scenario, suppose the total number of training functions is $N$, and the number of local clients is $K$. Each local client has $N_k = \frac{N}{K}$ functions generated from various functional spaces. We choose the Chebyshev polynomial space as the generating space for both training and testing functions, defined by 
$$p(x) = \sum_{i=0}^{M} a_i T_i(x), \qquad a_i \in [-1,1],$$
where $T_i$ is the Chebyshev polynomial of the first kind. The number of terms $M$ is set to 10 to avoid the computational cost associated with high-order polynomials.

The non-iid level across different functional spaces is controlled by varying the number of nonzero basis functions in the Chebyshev polynomial. For the case of two clients, one client is assigned the first $n$ nonzero basis functions (i.e., forward direction), while the other client is assigned the last $n$ nonzero basis functions (i.e., inverse direction):
\begin{align*}
    \text{Client 1: } &\mathcal{D} \sim \mathcal{P}_{\text{Chebyshev}}(n,\text{forward}) = \sum_{i=0}^{n} a_i T_i(x), \\
    \text{Client 2: } &\mathcal{D} \sim \mathcal{P}_{\text{Chebyshev}}(n,\text{inverse}) = \sum_{i=M-n-1}^{M-1} a_i T_i(x),
\end{align*}
where $n \in [1,10]$. This method controls the non-iid level by setting different supports for the Chebyshev polynomials, increasing the iid level as $n$ increases. The rationale is that functions generated from Chebyshev spaces with larger $n$ values will more closely approximate functions generated from the full Chebyshev space (Fig.~\ref{fig:assign}E). Thus, in the two-client scenario—where one client's data is generated using the forward method and the other using the inverse method—the data distributions on each client exhibit greater similarity as $n$ increases, validating the effectiveness of this non-iid control method for operator learning.

For three clients, the first client is supported by the first $n$ basis functions, the second client by the middle $n$ basis functions, and the third client by the last $n$ basis functions:
\begin{align*}
    \text{Client 1: } &\mathcal{D} \sim \mathcal{P}_{\text{Chebyshev}}(n,\text{forward}) = \sum_{i=0}^{n} a_i T_i(x),\\
    \text{Client 2: } &\mathcal{D} \sim \mathcal{P}_{\text{Chebyshev}}(n,\text{middle}) = \sum_{i=\lfloor\frac{M-n}{2}\rfloor}^{\lfloor\frac{M+n}{2}\rfloor} a_i T_i(x),\\
    \text{Client 3: } &\mathcal{D} \sim \mathcal{P}_{\text{Chebyshev}}(n,\text{inverse}) = \sum_{i=M-n-1}^{M-1} a_i T_i(x).
\end{align*}

%% file: content/Theory.tex
\section{Theory}
\label{sec:theory}
This section aims to quantify the differences between federated learning and traditional centralized learning. To achieve this, we first introduce the general definition of weight divergence, followed by a specification of the relative and absolute weight divergence used to measure the differences between the centralized and federated models (Sec.~\ref{subsec:wd}). Subsequently, we present the growth bound of weight divergence for the federated model under several assumptions and a detailed analysis (Sec.~\ref{subsec:growth_bound}).

\subsection{Weight divergence}
\label{subsec:wd}
Weight divergence in federated learning measures the relative difference between the weights learned by the centralized and federated optimization algorithms. First introduced in Ref.~\cite{Federated_noniid}, weight divergence is strongly associated with data heterogeneity. The definition of weight divergence, denoted as $\mathcal{E}_{WD}$, is as follows:
\begin{equation*}
    \mathcal{E}_{WD} \coloneqq \frac{\left\Vert \theta^{\mathcal{NN}_1}- \theta^{\mathcal{NN}_0}\right\Vert}{\left\Vert \theta^{\mathcal{NN}_0} \right\Vert},
\end{equation*}
where $\theta^{\mathcal{NN}_0}$ and $\theta^{\mathcal{NN}_1}$ represent the weights of two neural networks. We emphasize that the term ``relative weight divergence'' better fits this definition. Therefore, in this section, we will use the term weight divergence to refer to the absolute weight divergence, which is defined as
\begin{equation*}
    \mathcal{E}_{WD} \coloneqq \left\Vert \theta^{\mathcal{NN}_1} - \theta^{\mathcal{NN}_0} \right\Vert.
\end{equation*}
In this paper, we take $\mathcal{NN}_1$ as the federated model updated by FedAvg and $\mathcal{NN}_0$ as the model updated by centralized SGD. The weight divergence comparing the difference between FedAvg and SGD is thus defined as
\begin{equation*}
    \mathcal{E}_{WD} \coloneqq \left\Vert \theta^{\text{FedAvg}} - \theta^{\text{SGD}} \right\Vert.
\end{equation*}
Given the multi-round nature of federated optimization, we introduce the notation $\mathcal{E}_{WD}^{l, E}$ to represent the weight divergence between the federated model at the $l$-th global epoch and the $E$-th local epoch, and the centralized model at the $l \times E$-th epoch. This ensures a fair comparison between the federated and centralized models.

\begin{figure}[htbp]
    \centering
    \includegraphics[width=0.8\textwidth]{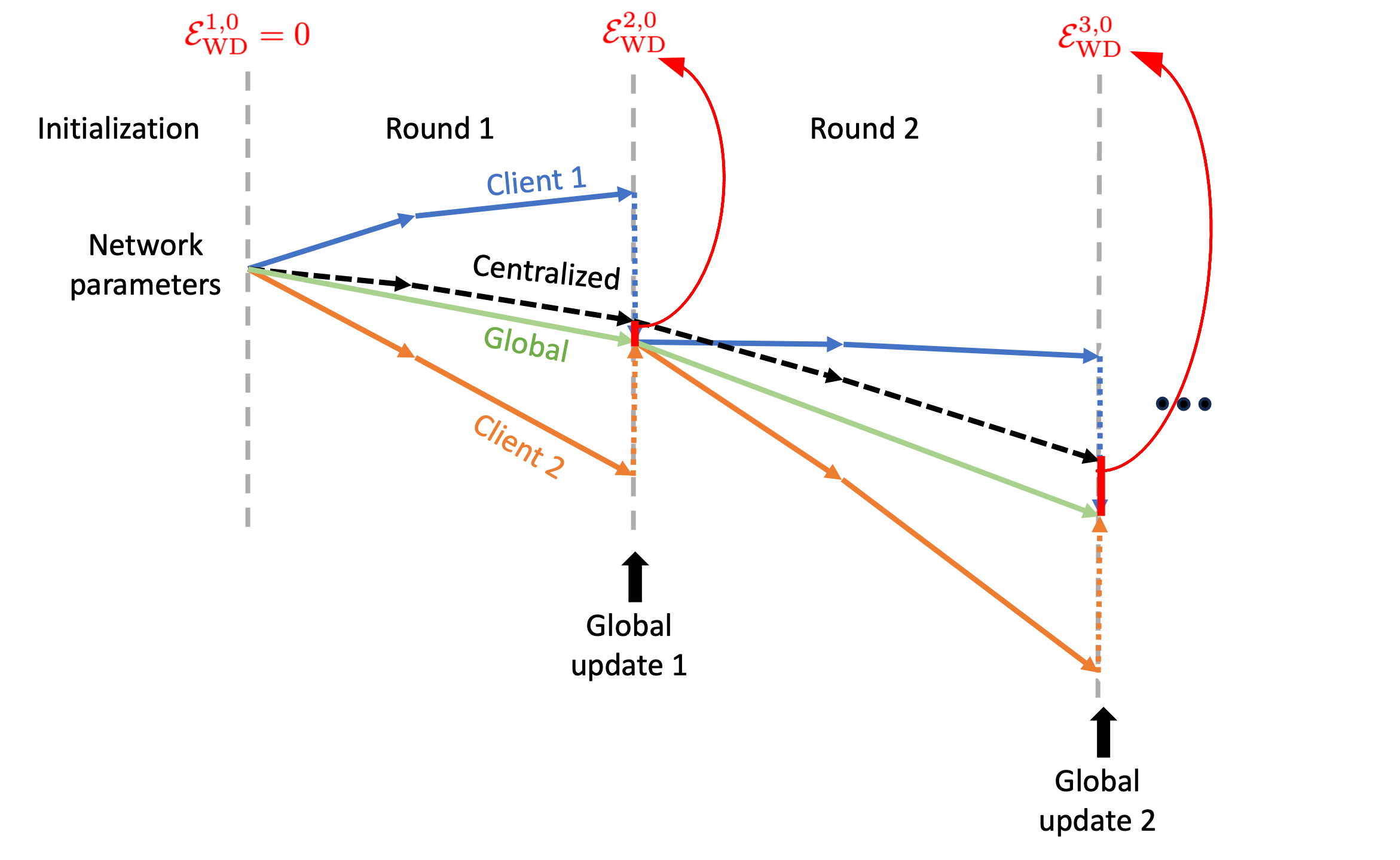}
    \caption{\textbf{Visualization of weight divergence in federated learning.} The black dashed line represents the gradient descent for the centralized model. The blue and orange lines correspond to the gradient descents for two clients in federated learning, while the green line depicts the gradient descent of the global model using the FedAvg algorithm.}
    \label{fig:vis_wd}
\end{figure}

Fig.~\ref{fig:vis_wd} provides a simple illustration of weight divergence during the updating process for two clients over two communication rounds. As shown in the figure, the neural network parameters are initialized at step 0. In subsequent steps, the global model is updated after two local gradient descent steps followed by one aggregation. The difference between the centralized model and FedAvg increases as more rounds of federated learning are performed.

Previous works have studied the convergence of federated learning~\cite{Federated_noniid,li2020federated,Li2020On}, PINNs~\cite{Shin_2020}, and DeepONets~\cite{deng2021convergence} separately under reasonable assumptions. Specifically, for the analysis of federated learning, weight divergence~\cite{Federated_noniid} is a critical factor that deteriorates the accuracy of federated models. Existing research has demonstrated that, after a finite number of synchronization rounds, weight divergence remains bounded~\cite{Federated_noniid}. FedProx~\cite{li2020federated}, an adapted federated aggregation algorithm incorporating weight divergence as a regularization term, has also proven highly effective in various federated learning tasks.

This work focuses on typical scientific machine learning scenarios and extends the discussion to derive an upper bound on the divergence between federated and centralized learning. Specifically, we will show in Sec.~\ref{subsec:growth_bound} that, under certain assumptions, after $E$ rounds of local updates, the weight divergence is linearly bounded by the local epoch $E$. Notably, the growth bound is independent of the data heterogeneity across federated clients.
\subsection{Growth bound of weight divergence}
\label{subsec:growth_bound}
We consider a scientific machine learning problem $\min \mathcal{L}(x)$ with dataset $D$. Our goal is to compare the performance of solving this problem using a centralized SGD algorithm for $E$ epochs and solving it under the federated learning scheme using the FedAvg algorithm for $E$ local epochs per global synchronization, both with the same learning rate $\eta$. The initialization of the two models is assumed to be identical. We assume full participation in federated learning with $K$ clients. Let $\{D_k\}_{k=1}^K$ represent the datasets of each client, where $D = \cup_{k=1}^{K}D_k$. Each client $k$ solves the problem $\mathcal{L}$ based on its local dataset $D_k$. Additionally, we impose the following bounded gradient assumption on the squared error loss $\ell$ of the neural network:

\begin{assumption}
    \label{assumption:bounded_grad}
    The gradient $\nabla\ell(\theta; d)$ is bounded during network training, i.e., there exists $M > 0$ such that $\forall \theta$ and $\forall d \in D$, $\|\nabla\ell(\theta,d)\| \leq M$.
\end{assumption}

 Given Assumption~\ref{assumption:bounded_grad}, we present the following theorem as the main result for the growth bound of the weight divergence.

\begin{theorem}
    \label{thm:temp}
    Consider a federated learning (FL) model with one global aggregation after $E$ local epochs and a learning rate $\eta$. Let $\mathcal{E}_{\text{WD}}^{1, E}$ denote the weight divergence of the FL model compared to a centralized model with the same initialization, learning rate, and trained for $E$ epochs. The dataset of the centralized model $D$ satisfies $D = \bigcup_k D_k$. Then, the weight divergence is bounded by a linear relation with $E$:
    $$\mathcal{E}_{\text{WD}}^{1,E}\leq 2\eta ME.$$
    The conclusion can also be extended to the $l$-th global epoch. The weight divergence $\mathcal{E}_{\text{WD}}^{l,E}$ between a federated model with $l$ global epoch and $E$ local epoch, and a centralized model with $l\times E$ epoch is bounded by
    $$\mathcal{E}_{\text{WD}}^{l,E}\leq 2\eta MEl.$$
\end{theorem}

%% file: content/FunctionApprox.tex
\section{Function approximation}
\label{sec:func_approx}
This section tests the federated neural networks in conducting function approximation tasks. For all following experiments, we use the Python library DeepXDE~\cite{lu2021deepxde} to implement the algorithms and train the neural networks. Codes for this study will be available in GitHub repository \url{https://github.com/lu-group/federated-sciml}.

We use two baselines to evaluate the performance of federated methods comprehensively. We use the centralized neural networks that are directly trained on the whole dataset as the first baseline model as a reference of lower error bound, and we use the extrapolation scenario where each client is trained only on local data without aggregation step during training as the reference of error upper bound. Despite minimal fluctuations, we anticipate the federated error to lie within the range of two baseline errors. With an increase in the level of iid data, we expect the errors of baseline models and federated methods to converge. The detailed hyperparameters used in training for all later experiments are summarized in Table~\ref{tab:arch}.

\subsection{1D function approximation}
\label{subsec:gramacy}
First, we consider the Gramacy \& Lee (2012) function~\cite{gramacy} as the one-dimensional example, which is defined as
\begin{equation*}
    f(x)=(x-1)^4+\frac{\sin(10\pi x)}{2x},\quad x\in[0.5,2.5].
\end{equation*}
We translate this function to $[-1,1]$ for later experiment and the normalized function becomes
\begin{equation*}
    f(x)=(x + 0.5)^4 -\frac{ \sin(10  \pi x) }{(2x+3)}.
\end{equation*}
In this example, to generate different levels of data heterogeneity, the number of partitions for each client varies from 1 to 50 over 200 uniformly distributed sample points within $[-1,1]$. We use the 1D data generation method for two clients in Fig.~\ref{fig:assign}A to build the dataset for both clients.

\begin{figure}[htbp]
    \centering
    \includegraphics[width=\textwidth]{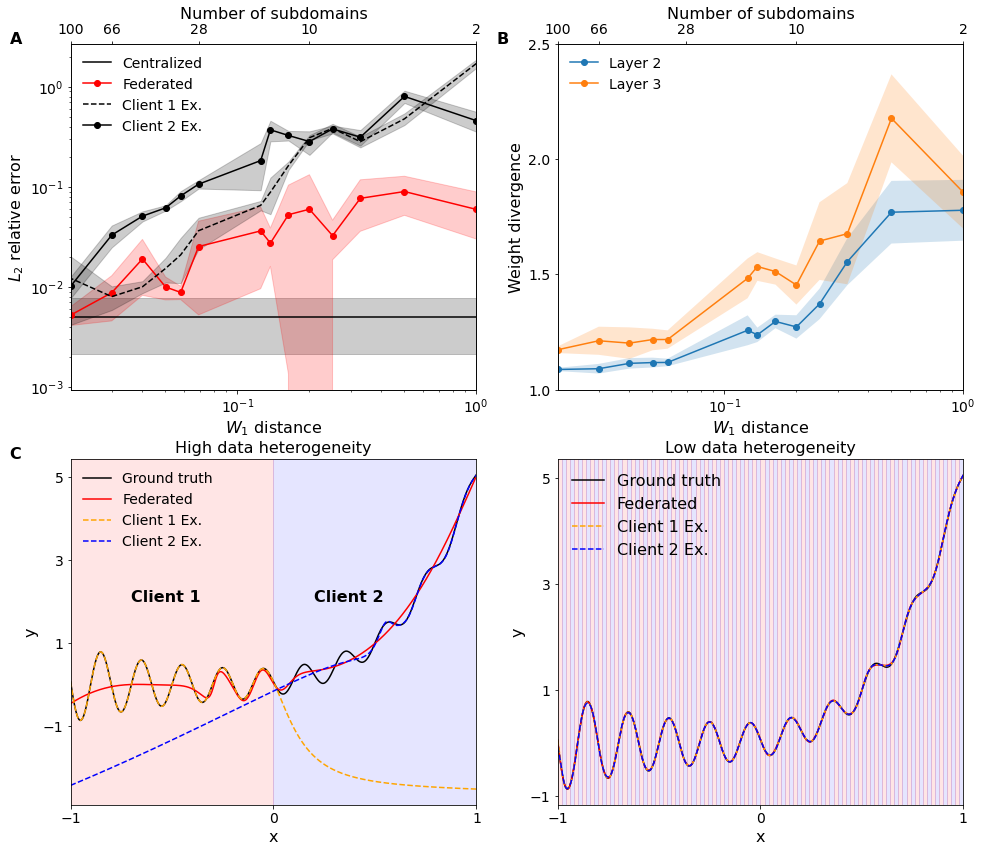}\caption{\textbf{Approximating the Gramacy \& Lee function in Section~\ref{subsec:gramacy}.} (\textbf{A}) $L_2$ relative error for different number of subdomains and $W_1$ distances. (\textbf{B}) Weight divergence of hidden layers for different numbers of subdomains and $W_1$ distances. (\textbf{C}) Examples include (left) 2 subdomains with high data heterogeneity and (right) 100 subdomains with low data heterogeneity. The shaded areas represent the subdomain partitions for two clients.}
    \label{fig:gramacy}
\end{figure}

When the value of $W_1$ decreases, the difference in the data distribution between two clients decreases, and the $L_2$ relative errors between the prediction results and actual values also decrease (Fig.~\ref{fig:gramacy}A). Apart from prediction errors, we also consider weight divergence, which is one of the factors that can impact the performance of federated learning on non-iid data. We found that the data heterogeneity between the two clients is positively correlated with the weight divergence between the neural networks of the two clients (Fig.~\ref{fig:gramacy}B).

Fig.~\ref{fig:gramacy}C are two examples of the approximated results of the Gramacy \& Lee function with high data heterogeneity and low heterogeneity scenarios. The data assigned to the two clients are represented by orange and blue shadow areas, respectively. In the extrapolation baselines, local models are trained only on local datasets without communication with the other, which implies that predicting the function on the whole domain is likely to fail. For example, for client 1 with left half training data points, the model performs good approximation on the left half function but falls on the right half function. On the contrary, the federated model can capture information about the whole domain via aggregating and broadcasting the parameters from local models. Moreover, when the data distributions of local clients are almost identical (Fig.~\ref{fig:gramacy}C Right), the performance of the federated model and extrapolating baselines is similar to the centralized model.

\subsection{2D function approximation}
\label{subsec:schaffer}
\begin{figure}[htbp]
    \centering
    \includegraphics[width=\textwidth]{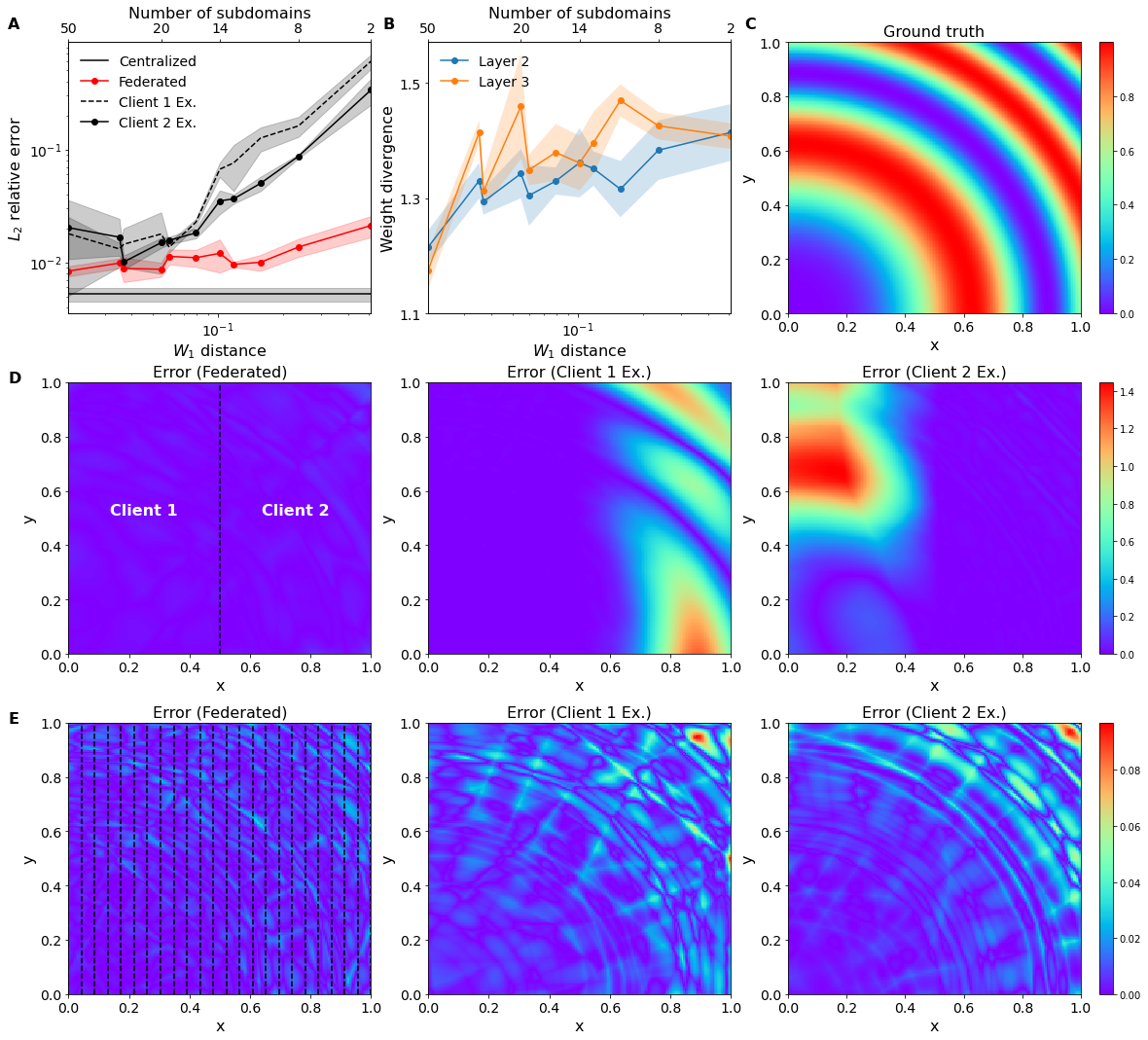}
    \caption{\textbf{Approximating the Schaffer function with two clients in Section~\ref{subsec:schaffer}}. (\textbf{A}) $L_2$ relative error for different number of subdomains and $W_1$ distances. (\textbf{B}) Weight divergence of hidden layers for different numbers of subdomains and $W_1$ distances. (\textbf{C}) The ground truth of the Schaffer function. (\textbf{D} and \textbf{E}) Comparison of
    federated model and two extrapolation baselines under (D) most non-iid and (E) most
    iid scenarios. The black dashed line visualizes the $x$-partition 2D data generation for two clients.}
    \label{fig:schaffer}
\end{figure}

Then we consider the Schaffer function in two-dimension~\cite{schaffer}, defined as 
$$f(x,y) = 0.5 + \frac{\sin^2(x^2-y^2)-0.5}{[1+0.001(x^2+y^2)]^2},\quad x\in[-1,1],\ y\in[-1,1].$$
The Schaffer function in 2D is symmetric, concerning both coordinates within the domain. Thus, to facilitate the generation of data heterogeneity, we consider an asymmetric subdomain of the function and select the right upper corner of the original function, i.e., we sample the data from the subdomain of $[0,1]\times[0,1]$ (Fig.~\ref{fig:schaffer}C). To generate different levels of data heterogeneity, we apply the $x$-partition 2D generation method ranges with the number of partitions ranging from 1 to 25. 

When the number of partitions increases and the value of $W_1$ decreases, the level of data heterogeneity decreases correspondingly and improves $L_2$ relative error from $10^{-1}$ to below $10^{-2}$ (Fig.~\ref{fig:schaffer}A). This overall trend between the $W_1$ and the performance of federated models is consistent with the 1D function approximation in Sec.~\ref{subsubsec:1d}. In contrast, improving approximation capability depends on the difficulty level of the target functions. Besides, the positive correlation between the data heterogeneity and the weight divergence also holds, except for some fluctuations (Fig.~\ref{fig:schaffer}B). Figs.~\ref{fig:schaffer}D and E are comparisons of the absolute error of prediction from federated model and extrapolating baselines with high and low data heterogeneity. When there is a lack of communication among clients, the missing information in the extrapolation region inevitably causes more significant errors, especially in the case of high non-iid levels. On the contrary, the federated model can take advantage of data on both clients with aggregation and broadcast, thereby performing better for both high and low data heterogeneity.

\begin{figure}[htbp]
    \centering
    \includegraphics[width=\textwidth]{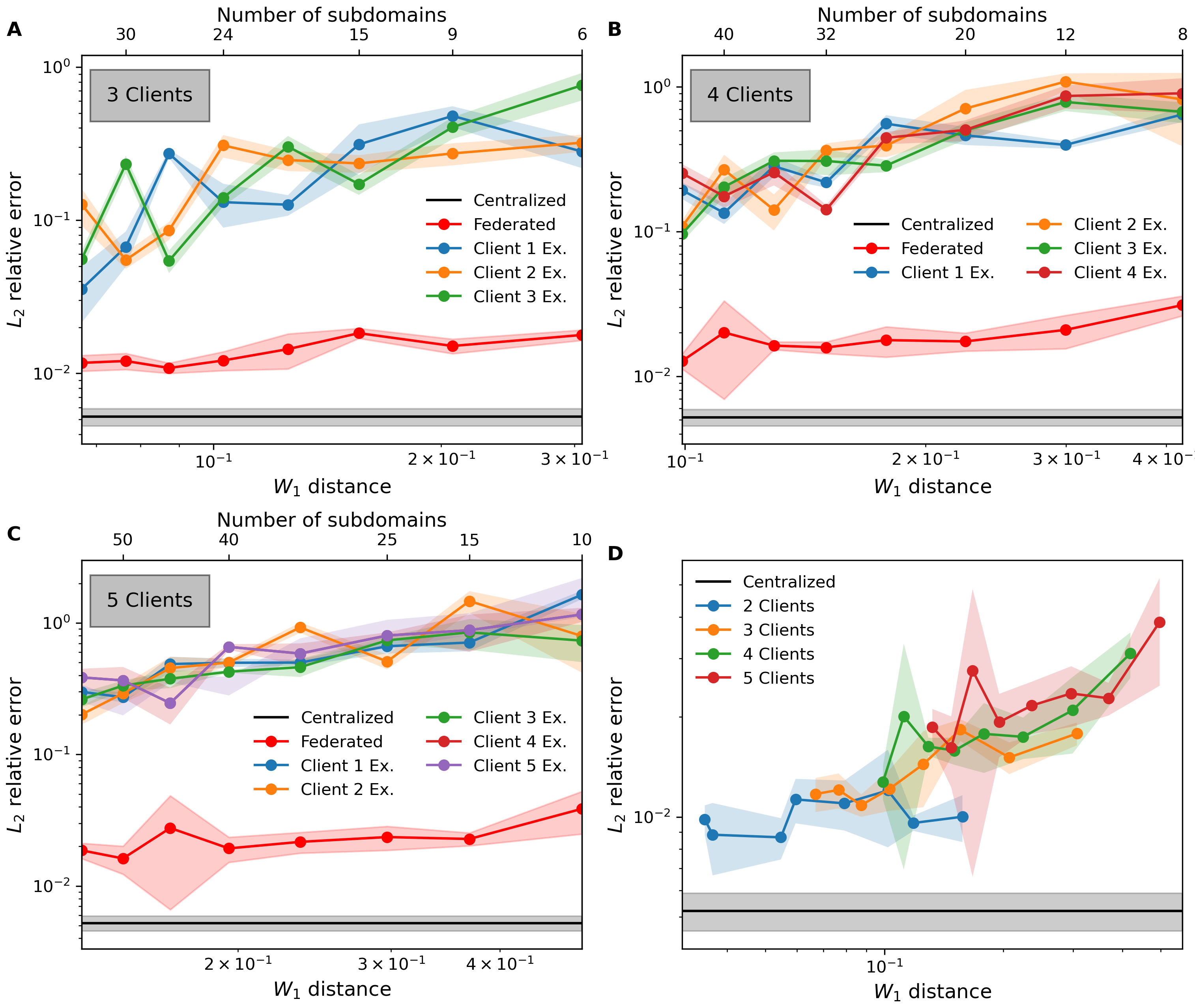}
    \caption{\textbf{Approximating the Schaffer function with multiple clients in Section~\ref{subsec:schaffer}.} (\textbf{A}--\textbf{C}) $L_2$ relative error for (A) three, (B) four, and (C) five clients for the different number of subdomains and mean pairwise $W_1$. (\textbf{D}) Comparison of the centralized baseline model and federated models with different numbers of clients.}
    \label{fig:schaffer_multi}
\end{figure}

Moreover, to explore the effect of the number of clients on federated learning, we trained the federated models with three, four, and five local clients to approximate the Schaffer function. As the direct computation of $W_1$ is not applicable for multiple clients scenario, we choose the mean pairwise $W_1$ defined in Eq.~\eqref{eq:pairw1} to quantify the data heterogeneity. For FedSciML with multiple clients, the trend that more considerable $W_1$ results in larger $L_2$ relative errors still holds, and the difference between extrapolating models and federated models becomes more prominent with more clients (Figs.~\ref{fig:schaffer_multi}A--C). We see that for multiple client scenarios, federated models and extrapolating baselines obtain more significant errors with increasing clients. As shown in Fig.~\ref{fig:schaffer_multi}D, the 2-client federated model has the most minor errors compared with the 5-client model. Besides, it is noted that the direct computation of $W_1$ for two clients and mean pairwise $W_1$ for multiclients inevitably give different $W_1$ ranges, and therefore, four error lines are not aligned in the figure. 

%% file: content/PINN.tex
\section{Physics-informed neural networks}
\label{sec:pinn}
In this section, after briefly introducing PINNs, we apply the proposed FedPINN to five examples and compare the performance with vanilla PINN and extrapolation baseline models. Examples include time-independent PDE, including 1D Poisson problem (Sec.~\ref{subsec:1dpoisson}) and 2D Helmholtz equation (Sec.~\ref{subsec:helmholtz}), time-dependent Allen-Cahn equation (Sec.~\ref{subsec:allencahn}), 2D inverse Navier-Stokes equation (Sec.~\ref{subsec:ns}), and inverse diffusion-reaction equation (Sec.~\ref{subsec:inverse_dr}). For the Allen-Cahn example, we also considered a multiclient scenario.

\subsection{Introduction of PINN}
PINN~\cite{karniadakis2021physics} is a class of deep learning algorithms that seamlessly integrate data and abstract mathematical operators, including PDEs with or without missing physics. This method uses prior physical knowledge to guarantee accuracy, interpretability, and robustness for predictions, even for extrapolatory and generalization tasks.

Consider a well-posed PDE defined on a bounded domain, 
\begin{equation*}
    \mathcal{F}[u](\mathbf{x}) = {f(\mathbf{x})}, \ {\mathbf{x}} \in \Omega,
\end{equation*}
with suitable boundary and initial conditions. We use a neural network $ \hat{u}(\theta)$ to represent the solution of the PDE. To embed the PDE into neural network training, we define one additional loss as 
\begin{equation*}
    \mathcal{L}_r(\theta; \mathbf{x}) = \|{\mathcal{F}[\hat{u}(\theta)]}(\mathbf{x})-{f}(\mathbf{x})\|_2^2.
\end{equation*}
In this study, we enforce the hard boundary constraints for all examples~\cite{lu2021physics}. Without specification, the default setting during training is hyperbolic tangent activation and Adam optimizer with a learning rate of 0.001. Table~\ref{tab:arch} summarizes the hyperparameters used to train neural networks.

In FedPINN, the entire computational domain is partitioned into many subdomains, and each client has the information of a single subdomain. This shares some similarities with the domain decomposition methods for solving PDEs, such as PINNs with domain decomposition~\cite{jagtap2020conservative, jagtap2021extended, shukla2021parallel}. However, FedPINN and domain decomposition methods are conceptually different. In domain decomposition, the partition of the computation domain must be known so that each subdomain knows its adjacent subdomains for information exchange, or there is a centralized server node to facilitate information exchange with all subdomains. However, in FedPINNs, knowledge of other local clients' domain locations is not required. In this federated learning framework, domain locations from local clients remain concealed.

\subsection{1D Poisson problem}
\label{subsec:1dpoisson}

We consider the following 1D Poisson problem: 
$$-\Delta u = \sum_{i=1}^{4} i \sin(ix) + 8 \sin(8x), \quad x \in [0, \pi],$$
with Dirichlet boundary conditions
$$u(0) = 0, \quad u(\pi) = \pi.$$
The exact solution is given by 
$$u(x) = x + \sum_{i=1}^4 \frac{\sin(ix)}{i} + \frac{\sin(8x)}{8}.$$
We sample 32 uniformly distributed points in the domain $[0, \pi]$ as collocation points for the training data and apply the 1D data generation method with the number of partitions $n \in [6, 8, 10, 16, 32]$ to simulate different levels of data heterogeneity.
\begin{figure}[htbp]
    \centering
    \includegraphics[width=\textwidth]{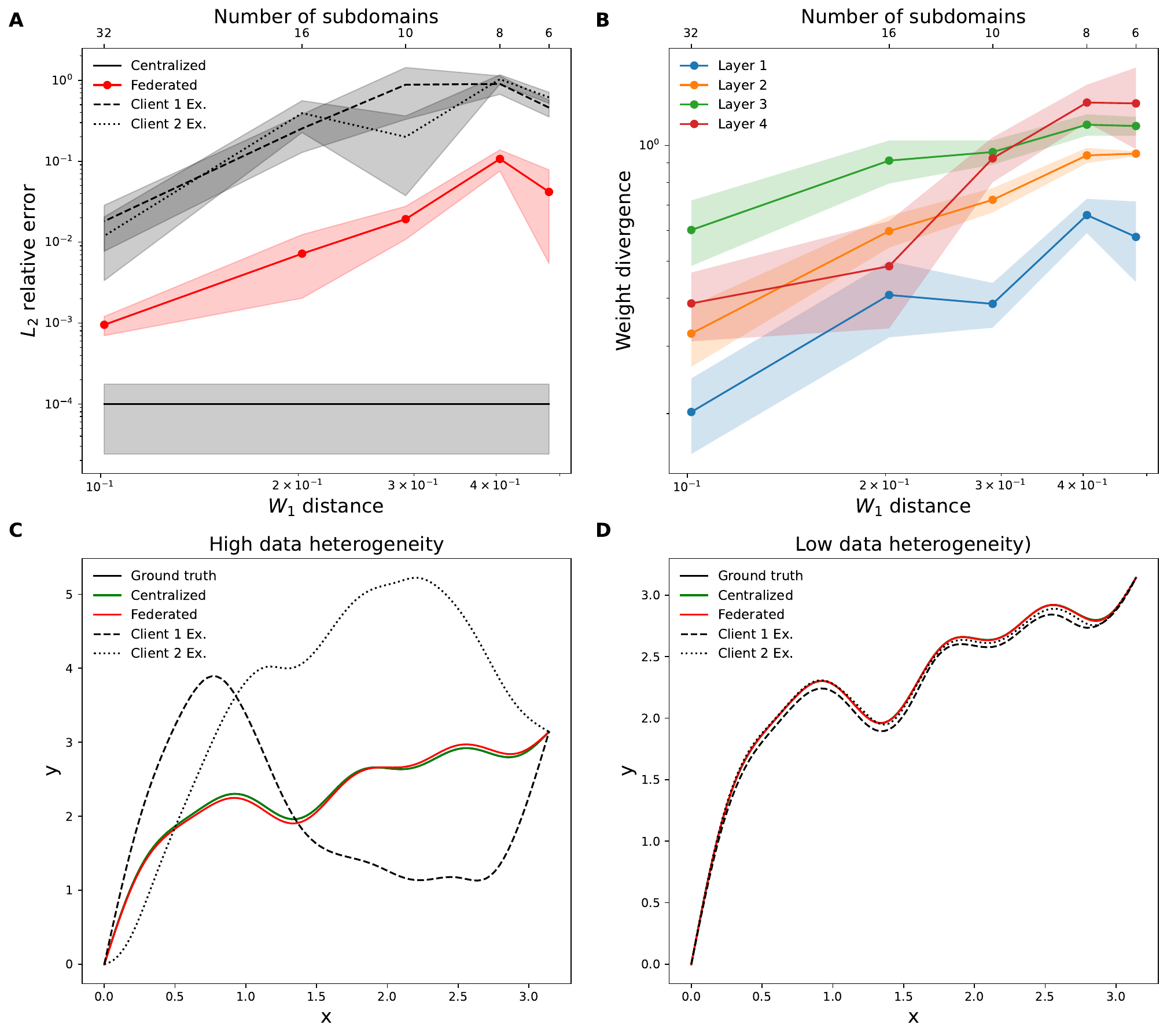}
    \caption{\textbf{Solving the 1D Poisson equation in Section~\ref{subsec:1dpoisson}.} (\textbf{A}) $L_2$ relative error for different numbers of subdomains and $W_1$ distances. (\textbf{B}) Weight divergence of hidden layers for different $W_1$ values. (\textbf{C} and \textbf{D}) Comparison of the federated model, baseline model, and two extrapolation models under (C) the most non-iid and (D) the most iid scenarios.}
    \label{fig:pinn_poisson}
\end{figure}

The accuracy achieved by both the extrapolation clients and the federated model decreases as the $W_1$ distance between data distributions increases (Fig.~\ref{fig:pinn_poisson}A). The federated model outperforms the extrapolation models, which perform similarly to the centralized baseline when the $W_1$ distance is small. The weight divergence across all four linear layers grows with increasing data heterogeneity (Fig.~\ref{fig:pinn_poisson}B). Fig.~\ref{fig:pinn_poisson}C and D are the predictive solutions of the federated, centralized baseline and the extrapolation models in the most non-iid and most iid settings. The federation of client models significantly improves solving the 1D Poisson system.

\subsection{2D Helmholtz equation}
\label{subsec:helmholtz}

Next, we consider the 2D Helmholtz equation over the square domain
\begin{equation*}
    - u_{xx}-u_{yy} - k_0^2 u = f, \qquad  \Omega = [0,1]^2,
\end{equation*}
with the Dirichlet boundary conditions $u(x,y)=0$ for $(x,y)\in \partial \Omega$ and source term $f(x,y)=k_0^2 \sin(k_0x)\sin(k_0y)$, where the wavenumber $k_0=2\pi n$ with $n=2$. We use the collocation points density of 12 points per wavelength for the training data along each direction, i.e., the number of sample points is 24. Besides, to simulate different levels of data heterogeneity, we apply the 2D $ xy$-partition with $n\in [2,4,6,10,12,24]$.

\begin{figure}[htbp]
    \centering
    \includegraphics[width=\textwidth]{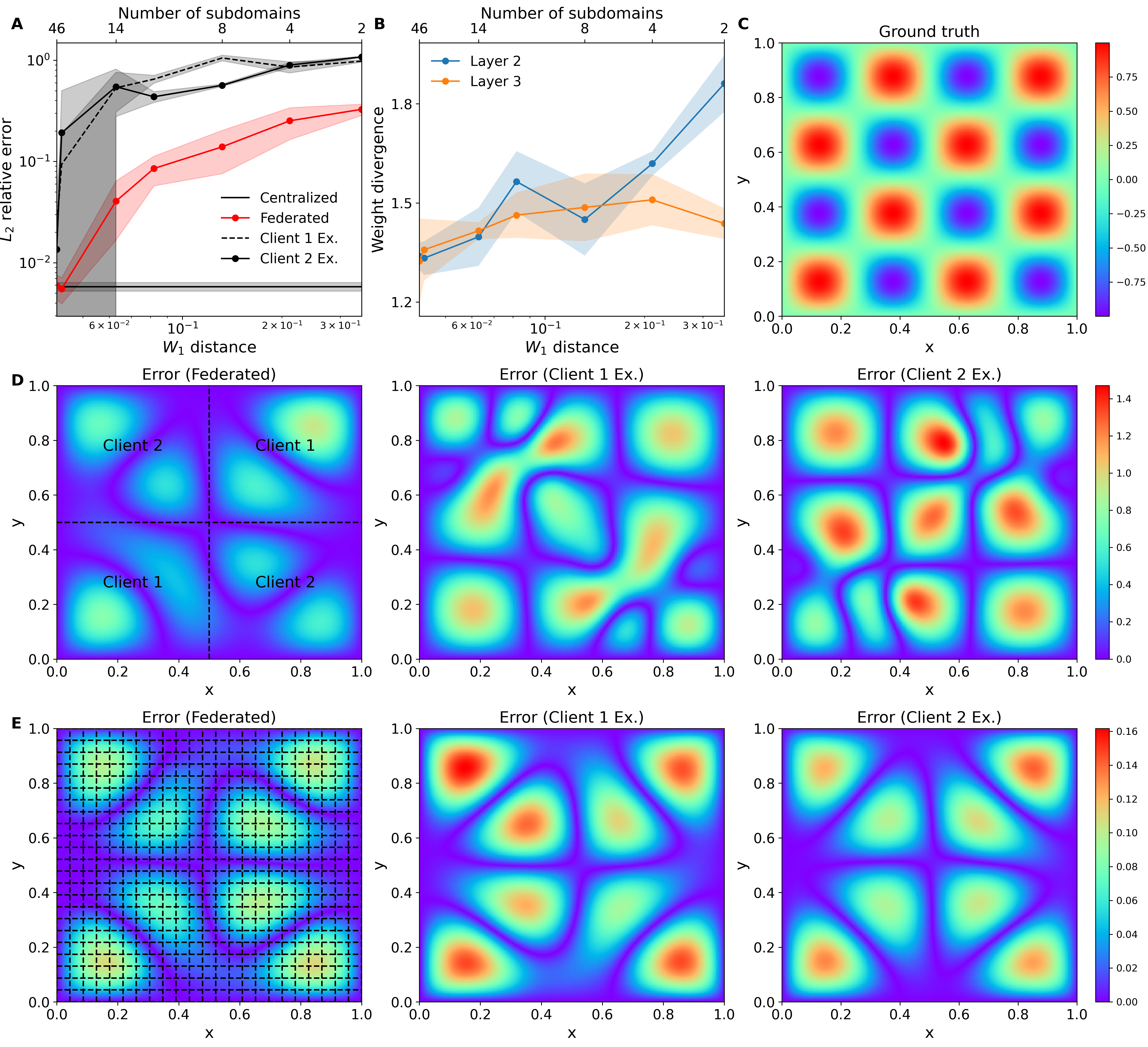}
    \caption{\textbf{Solving the 2D Helmholtz equation in Section~\ref{subsec:helmholtz}.} (\textbf{A}) $L_2$ relative error for different number of subdomains and $W_1$ distances. (\textbf{B}) Weight divergence of hidden layers for different numbers of subdomains and $W_1$ distances. (\textbf{C}) The ground truth solution. (\textbf{D} and \textbf{E}) Comparisons of
    federated model and two extrapolation baselines under (D) most non-iid and (E) most
    iid scenarios.}
    \label{fig:pinn_helmholtz}
\end{figure}

For FedPINN, the increasing $W_1$ corresponds to the increasing heterogeneity of two clients' data distribution, resulting in larger $L_2$ relative errors of the prediction results (Fig.~\ref{fig:pinn_helmholtz}A). As $W_1$ decreases with an increasing number of subdomains, data distribution of local clients converges to the original dataset, and both the federated model and extrapolating baselines have improved performance. Apart from the prediction error, we also observe that the weight divergence of the second and third layers of the FNN is approximately positively correlated with $W_1$ (Fig.~\ref{fig:pinn_helmholtz}B). However, the second layer is more sensitive than the third layer. Figs.~\ref{fig:pinn_helmholtz}D and E are examples of learned solutions in most non-iid and iid scenarios. In most non-iid cases (Fig.~\ref{fig:pinn_helmholtz}D), the data blocks generated by the $xy$-partition method are distinguished by the black dashed line, where the first client takes the bottom left and upper right blocks, and the second client takes the rest. In the extrapolation baselines, local models are trained only on local datasets without communication with the other, which demonstrates the weak performance of learning the solution in the extrapolating regions. For instance, for client 1, the model has a more significant absolute error in the bottom right and upper left areas, where it lacks information on these regions during training. The federated model, however, can aggregate the overall information within the domain well by efficient communication between local models.

\subsection{Allen-Cahn equation}
\label{subsec:allencahn}
We consider solving the Allen-Cahn equation, governed by
\begin{equation*}
    \frac{\partial u}{\partial t} = d\frac{\partial^2 u}{\partial x^2} + 5(u - u^3), \quad x \in [-1, 1], \quad t \in [0, 1],
\end{equation*}
with the initial condition $u(x, 0) = x^2 \cos(\pi x)$ and boundary conditions
$u(-1, t) = u(1, t) = -1,$
where $d$ is 0.001. We sample 8000 residual points within the domain and 400 points on the boundary for the training data. Additionally, we sample 800 residual points for the initial condition during training. Since the Allen-Cahn equation is time-dependent, rather than partitioning both the spatial and temporal domains, we consider only the partitioning of the spatial domain and apply the 1D data partition method with the number of subdomains $n \in [2, 4, 6, 8, 10, 20, 32, 40]$. For the multiclient experiment, we consider three clients using the same partitioning method but with different numbers of subdomains, $n \in [3, 6, 9, 12, 15, 18, 30, 60]$, ensuring that all clients retain complete temporal information over their assigned spatial domain. The number of training epochs, neural network structure, activation function, and learning rate are consistent across both the two-client and multiclient scenarios.

\begin{figure}[htbp]
    \centering
    \includegraphics[width=\textwidth]{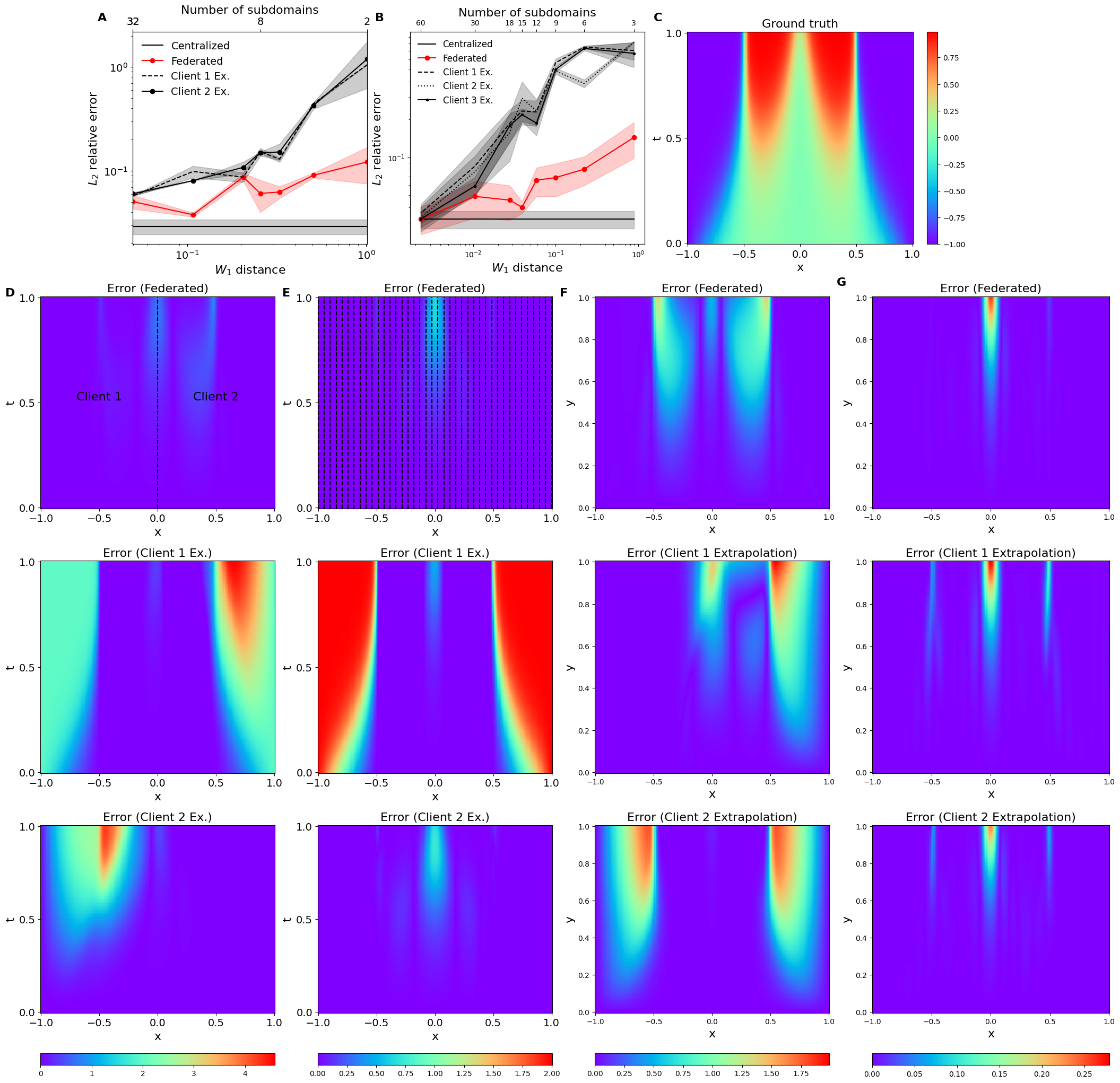}
    \caption{\textbf{Solving the Allen-Cahn equation with two and three clients in Section~\ref{subsec:allencahn}.} (\textbf{A}) $L_2$ relative error for different numbers of subdomains and $W_1$ distances under the two-client setting. (\textbf{B}) $L_2$ relative error for different numbers of subdomains and $W_1$ distances under the multiclient setting. (\textbf{C}) Ground truth solution. (\textbf{D} and \textbf{E}) Comparison of federated model and two extrapolation baselines under most non-iid and most iid scenarios in the two-client setting. (\textbf{F} and \textbf{G}) Comparison of federated model and two extrapolation baselines under most non-iid and most iid scenarios in the multiclient setting.}
    \label{fig:pinn_allencahn}
\end{figure}

The relationship between $L_2$ relative errors and the $W_1$ distance is consistent with the previous time-independent example (Figs.~\ref{fig:pinn_allencahn}A and B). The FedPINN outperforms the baseline PINN in extrapolation scenarios, with both models improving as the level of iid increases. Higher levels of data heterogeneity between clients correspond to larger $L_2$ relative errors, both in the two-client and multiclient settings, where the $W_1$ distance for the multiclient setting is computed using the mean pairwise $W_1$, as defined in Eq.~\eqref{eq:pairw1}.

Fig.~\ref{fig:pinn_allencahn}C shows the visualization of the ground truth solution, while Figs.~\ref{fig:pinn_allencahn}D and E present two examples of the learned solutions in the most non-iid and most iid scenarios under the two-client setting. Both clients have complete temporal information in this $x$-partition data generation method but differ only in spatial regions. For example, in most non-iid cases, the first client handles data points with $x \in [-1, 0]$ and $t \in [0, 1]$, while the second client handles $x \in [0, 1]$ and $t \in [0, 1]$. Figs.~\ref{fig:pinn_allencahn}F and G display the absolute error of two extrapolation clients alongside the error of the corresponding federated server. Notably, federated learning significantly improves the performance of extrapolation clients, regardless of the level of data heterogeneity.

\subsection{Inverse problem of the 2D Navier-Stokes equation}
\label{subsec:ns}
We consider the inverse problem of the given Navier-Stokes equation of incompressible flow around the cylinder with Re $=100$. The governing equations are 
\begin{equation*}
    u_t + C_1 (uu_x+vu_y) = -p_x + C_2(u_{xx} + u_{yy}) ,
\end{equation*}
\begin{equation*}
    v_t + C_1 (uv_x + vv_y) = - p_y + C_2 (v_{xx} + v_{yy}),
\end{equation*}
where $u,v$ are $x$ and $y$ components of the velocity field, $p$ is the pressure, $C_1, C_2$ are unknown parameters. Here, we aim to infer the values of $C_1$ and $C_2$ from the dataset of $u,v$ and $p$. The ground truth of $C_1$ and $C_2$ are 1 and 0.01. The geometry domain of interest is $[1,8]\times [-2,2]$ and the time domain is $[0,1]$. We use 200 boundary points and 100 initial points. We use 700 in-domain mesh points additionally. We partition the spatial domain in this problem using the 2D $x$-partition method in Fig.~\ref{fig:assign}A, with the number of subdomains $n\in[2,4,6,8,10,20]$. 

\begin{figure}[htbp]
    \centering
    \centerline{\includegraphics[width = \textwidth]{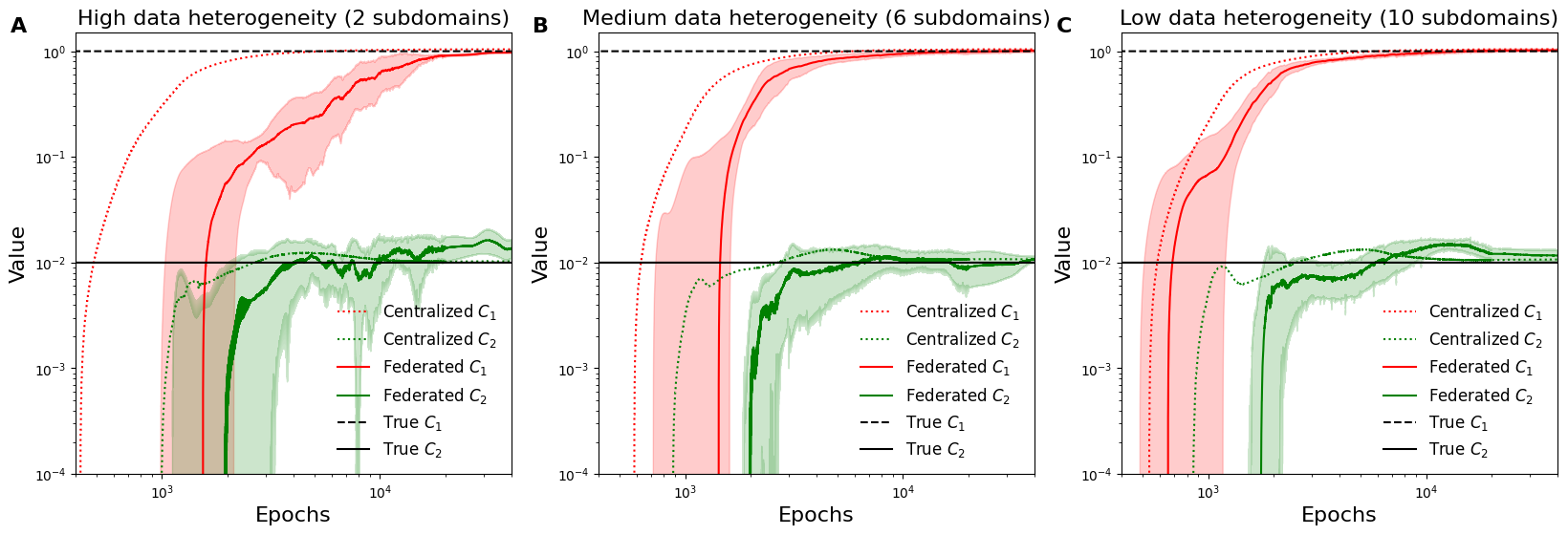}}
    \caption{\textbf{Solving the inverse Navier-Stokes problem with different data heterogeneity in Section \ref{subsec:ns}.} The predicted values of $C_1$ and $C_2$ converge to the true values. (\textbf{A}) The spatial domain is partitioned into two uniform subdomains. (\textbf{B}) Six subdomains. (\textbf{C}) Ten subdomains.}
    \label{fig:2DNS}
\end{figure}

For this inverse problem, we average the gradients from clients in the aggregation step and the inferred values in the aggregation step. Hence, after a few iterations, the two clients will have a similar learned value for the variables of interest. Fig.~\ref{fig:2DNS} is the trend of predicted $C_1$ and $C_2$ values versus the training epochs. We also record the minimum number of epochs the federated model needs to achieve the required accuracy on the two variables $C_1$ and $C_2$ in Table.~\ref{tab:ns}. In general, achieving better accuracy requires more epochs for all three settings. Specifically, less data heterogeneity, i.e., more partitions, will lead to faster model convergence. We conclude that data heterogeneity negatively correlates with the convergence rate and the performance of the federated inverse problem. In Fig.~\ref{fig:2DNS}A, the standard deviation of both unknown variables is much larger than the standard deviation of unknown variables in Figs.~\ref{fig:2DNS}B and C. This offers strong evidence that data heterogeneity also relates to the stability of the federated model. Less data heterogeneity would lead to a more stable federated model.

\begin{table}[htbp]
\centering
\caption{\textbf{Comparison of computational costs of the 2D inverse Navier-Stokes equation in Section \ref{subsec:ns}.} Data in the table represents the least epochs for the target variable to achieve the error threshold during the learning process. Data heterogeneity varies by columns, and error thresholds vary by rows.}
\begin{tabular}{@{}ccccccccc@{}}
\toprule
\multirow{2}{*}{Error threshold} & \multicolumn{2}{c}{2 subdomains} & \multicolumn{2}{c}{6 subdomains} & \multicolumn{2}{c}{10 subdomains} & \multicolumn{2}{c}{Centralized} \\ \cmidrule(l){2-9} 
      & $C_1$ & $C_2$ & $C_1$ & $C_2$ & $C_1$ & $C_2$ & $C_1$ & $C_2$ \\ \midrule
10\%  & 18k & 3.4k  & 6.8k  & 4.5k  & 5.5k  & 5.3k  & 2.9k  & 1k  \\
5\%    & 28k & 3.5k  & 10.2k & 5.3k  & 8.3k  & 5.5k  & 3.8k  & 2.3k  \\
1\%    & --    & --    & 23.6k & 5.7k  & 13k   & 5.7k  & 5.3k  & 2.4k  \\
0.5\%  & --    & --    & 25.9k & 5.7k  & 13.8k & 5.7k  & 5.6k  & 2.4k  
  \\ \bottomrule
\end{tabular}

\label{tab:ns}
\end{table}

\subsection{Inverse problem of the diffusion-reaction system}
\label{subsec:inverse_dr}
To further evaluate the performance of the proposed FedPINN in the inverse problem, we consider inferring the space-dependent reaction rate $k(x)$ in the one-dimensional diffusion-reaction system
$$\lambda\frac{\partial^2u}{\partial x^2}-k(x)u=f,\qquad x\in[0,1],$$
with zero boundary condition 
$u(x)=0\text{ at }x=0 \text{ and }x=1$. The diffusion coefficient $\lambda$ is 0.01, and $f=\sin(2\pi x)$ is the source term. The
objective is to infer $k(x)$ given measurements on the solute concentration $u$. The exact solution of this problem is
$$k(x)=0.1+\exp\left[-0.5\frac{(x-0.5)^2}{0.15^2}\right].$$
For this inverse problem, given that the unknown $k$ is a function of $x$ instead of a constant number, we train another network to approximate $k$ apart from the neural network for $u$. We use 24 observations of $u$ and 10 PDE residual points for training. In contrast, in the federated setting, the observation points are partitioned and assigned to different clients based on the 1D generation method in Section~\ref{sec:method} with $n\in [2, 4, 6, 8, 12, 24]$.

\begin{figure}[htbp]
    \centering
    \includegraphics[width=\textwidth]{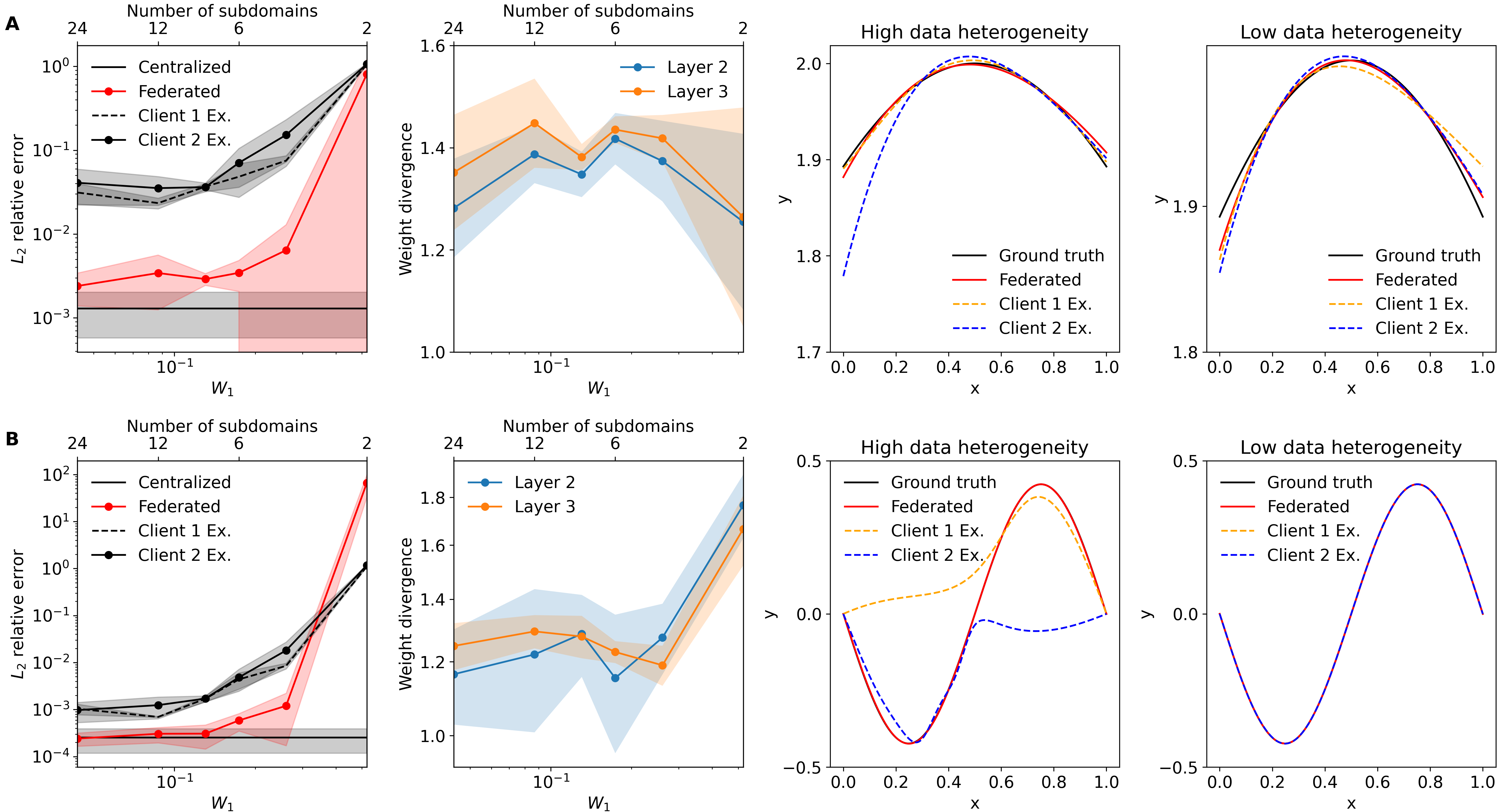}
    \caption{\textbf{Solving the inverse diffusion-reaction problem in Section~\ref{subsec:inverse_dr}.} (\textbf{A}) Results of the unknown function $k(x)$. (\textbf{B}) Results of the PDE solution $u(x)$. (First column) $L_2$ relative error for different numbers of subdomains and $W_1$ distances. (Second column) The weight divergence of hidden layers for different numbers of subdomains and $W_1$ distances. (Third column) The federated model under the most non-iid setting. (Fourth column) The federated model under the most iid setting.}
    \label{fig:pinn_inverse_dr}
\end{figure}

For the inverse problem, the $L_2$ relative errors for both $k(x)$ and $u(x)$ exhibit an increasing trend concerning the larger $W_1$ (the first column of Figs.~\ref{fig:pinn_inverse_dr}A and B), which is similar to the forward problem. However, no clear trend of weight divergence is observed in inferring the unknown $k(x)$ (the second column of Fig.~\ref{fig:pinn_inverse_dr}A). The last two columns present two examples of inferred $k(x)$ and learned solution $u(x)$ with high and low data heterogeneity.

%% file: content/OperatorLearning.tex
\section{Operator learning}
\label{sec:deeponet}

In this section, after briefly explaining the setting of operator learning and the architecture of DeepONets, we test the performance of the proposed FedDeepONet with antiderivative equation (Sec.~\ref{subsec:ad}), diffusion-reaction equation (Sec.~\ref{subsec:dr}), and Burgers' equation (Sec.~\ref{subsec:burgers}). Moreover, we explore the performance of FedDeepONet for different numbers of local iterations and find the insensitivity to communication efficiency under operator learning scenario (Sec.~\ref{subsec: fed_deeponet_insensitivity}). 

\subsection{Introduction of DeepONet}
DeepONet~\cite{deeponetNatureML} was developed to learn operators based on the universal approximation theorem of neural networks for operators. Consider two function spaces $\mathcal{V}$ and $\mathcal{U}$, where a function $v\in\mathcal{V}$ is defined on the domain $D \subset \mathbb{R}^d$:
$$
v: D \ni x \mapsto v(x) \in \mathbb{R},
$$
and a function $u\in\mathcal{U}$ defined on the domain $D'\subset \mathbb{R}^{d'} $:
$$
u: D' \ni \xi \mapsto u(\xi) \in \mathbb{R}.
$$
Let $\mathcal{G}$ be an operator that maps $\mathcal{V}$ to $\mathcal{U}$:
$$
\mathcal{G}: \mathcal{V} \to \mathcal{U}, \quad v \mapsto u.
$$
DeepONet can be used to learn the operator $\mathcal{G}$ with two sub-networks: a trunk network and a branch network. For $m$ scattered locations $\{x_1, x_2, \dots, x_m\}$ in $D$, the branch network takes the function evaluations $[v(x_1), v(x_2), \dots, v(x_m)]$ as the input, and the output of the branch network is $[b_1(v), b_2(v), \dots, b_p(v)]$, where $p$ is the number of neurons. The trunk network takes $\xi$ as the input and the outputs $[t_1(\xi), t_2(\xi), \dots, t_p(\xi)]$. Then, by taking the inner product of trunk and branch outputs, the output of DeepONet is
$$
\mathcal{G}(v)(\xi) = \sum_{k=1}^p b_k(v)t_k(\xi) + b_0,
$$
where $b_0 \in \mathbb{R}$ is a bias.

\subsection{Antiderivative operator}
\label{subsec:ad}
The first example is the ordinary differential equation defined by
\begin{equation*}
    \frac{du(x)}{dx} = v(x), \quad x\in [0,1],
\end{equation*}
with an initial condition $u(0)=0$. We use the DeepONet to learn the solution operator, i.e., the antiderivative operator 
$$\mathcal{G}: v(x) \mapsto u(x)=\int_0^x v(\tau)d\tau.$$
The input functions $v$ are sampled from the Chebyshev polynomial space defined in Sec.~\ref{subsubsec:operator}. For training datasets, different directions are chosen for two clients with the same number of basis functions, i.e., $\Omega_1\sim \mathcal{P}_{\text{Chebyshev}}(n,\text{forward})$, $\Omega_2\sim \mathcal{P}_{\text{Chebyshev}}(n,\text{inverse})$ with $n\in[1,10]$. Centralized and extrapolating models are trained with 200 functions, while the federated model is trained with 200 functions in total, where each client has 100 functions locally. After training, all models are tested on 1000 functions sampled from the entire Chebyshev polynomial space with $M=10$. The reference solution $u(x)$ of the ODE is obtained by Runge-Kutta (4,5).

\begin{figure}[htbp]
    \centering
    \includegraphics[width=\textwidth]{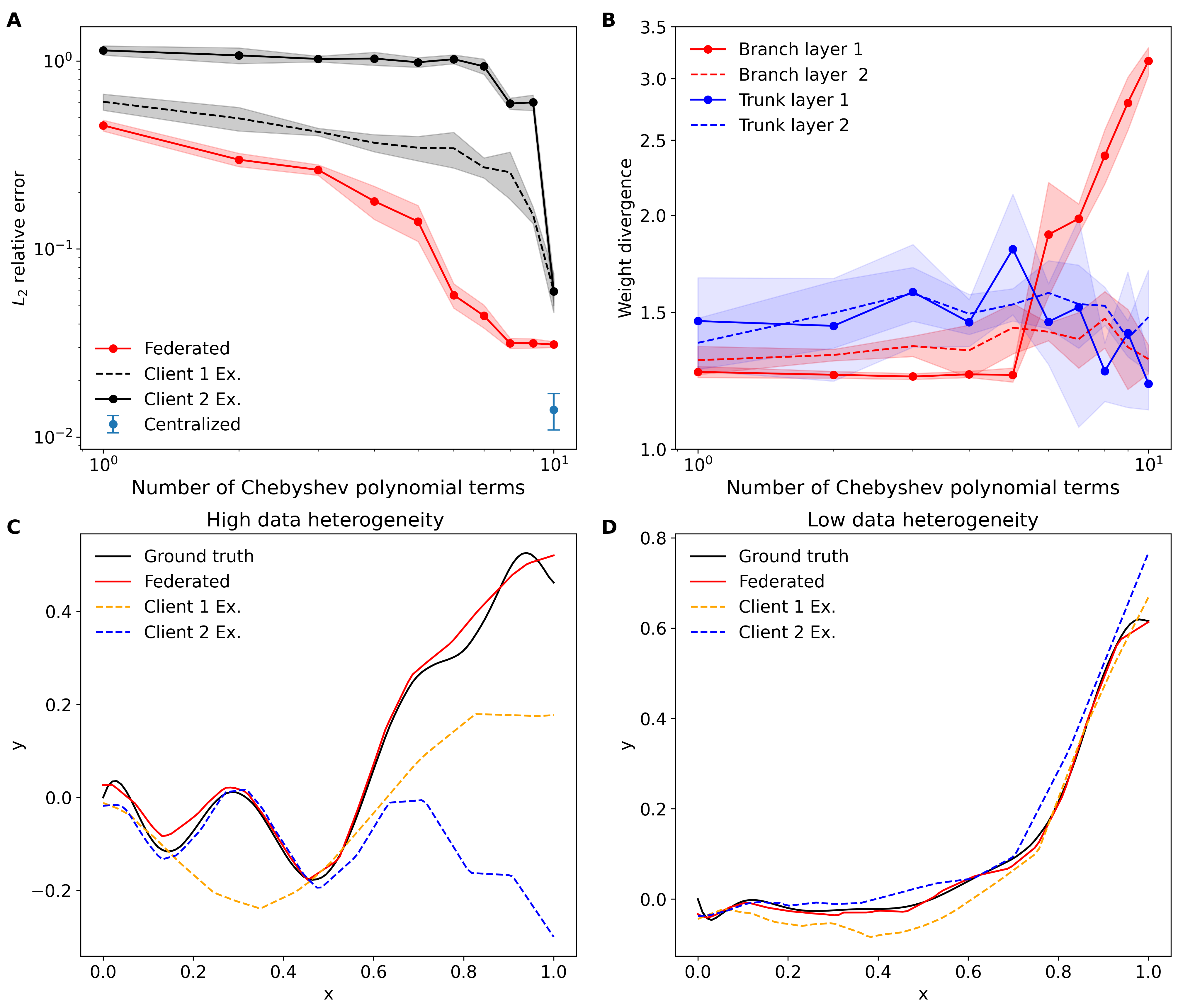}
    \caption{\textbf{Learning antiderivative operator in Section~\ref{subsec:ad}.} (\textbf{A}) $L_2$ relative error for different numbers of nonzero Chebyshev polynomial terms. (\textbf{B}) Weight divergence of hidden layers with respect to different numbers of nonzero Chebyshev polynomial terms. (\textbf{C} and \textbf{D}) Comparison of
    federated model and two extrapolation baselines for the number of nonzero Chebyshev polynomial terms (C) $n=6$ and (D) $n=10$.}
    \label{fig:deeponet_ad}
\end{figure}

In a federated operator learning setting, the data heterogeneity decreases as the number of nonzero terms increases, which further results in the decrease of $L_2$ relative error (Fig.~\ref{fig:deeponet_ad}A). Moreover, the performance of FedDeepONet exceeds the two extrapolating baselines while approximating the centralized model with diminishing data heterogeneity. It is noted that in Fig.~\ref{fig:deeponet_ad}A there is a large improvement in error from $10^{-1}$ to $5\times 10^{-2}$ when $n$ increases from 5 to 6. This may result from the fact that the total knowledge of the FedDeepONet, despite being distributed to two clients, can cover the full Chebyshev space. Another finding is that in the operator learning setting, the weight divergence for both trunk and branch layers is relatively less sensitive to the non-iid level change than approximating functions or learning a specific solution (Fig.~\ref{fig:deeponet_ad}B). Figs.~\ref{fig:deeponet_ad}C and D are two examples of the performance of given models with high data heterogeneity ($n=6$) and low data heterogeneity ($n=10$). As the extrapolation models are less satisfactory, the FedDeepONet can utilize the information on both clients without directly communicating local data and better perform in learning the antiderivative operator.

\subsection{Diffusion-reaction equation}
\label{subsec:dr}

The second example is the diffusion-reaction equation defined by
\begin{equation*}
    \frac{\partial u}{\partial t} = D \frac{\partial^2 u}{\partial x^2} + ku^2 + v(x), \quad x \in [0,1], t \in [0,1],
\end{equation*}
with zero initial and boundary conditions. In this example, $k$ and $D$ are set at 0.01. DeepONet is trained to learn the mapping from the source term $v(x)$ to the solution $u(x, t)$:
$$\mathcal{G}: v(x) \mapsto u(x,t).$$
Models are trained with 500 functions, where each client has 250 functions locally, and tested on 1000 functions. The training and testing of Chebyshev functional spaces follow the same setting as in Sec.~\ref{subsec:ad}. The reference solution $u(x,t)$ is obtained by a second-order finite difference method with a mesh size of $101\times 101$.

\begin{figure}[htbp]
    \centering
    \includegraphics[width=\textwidth]{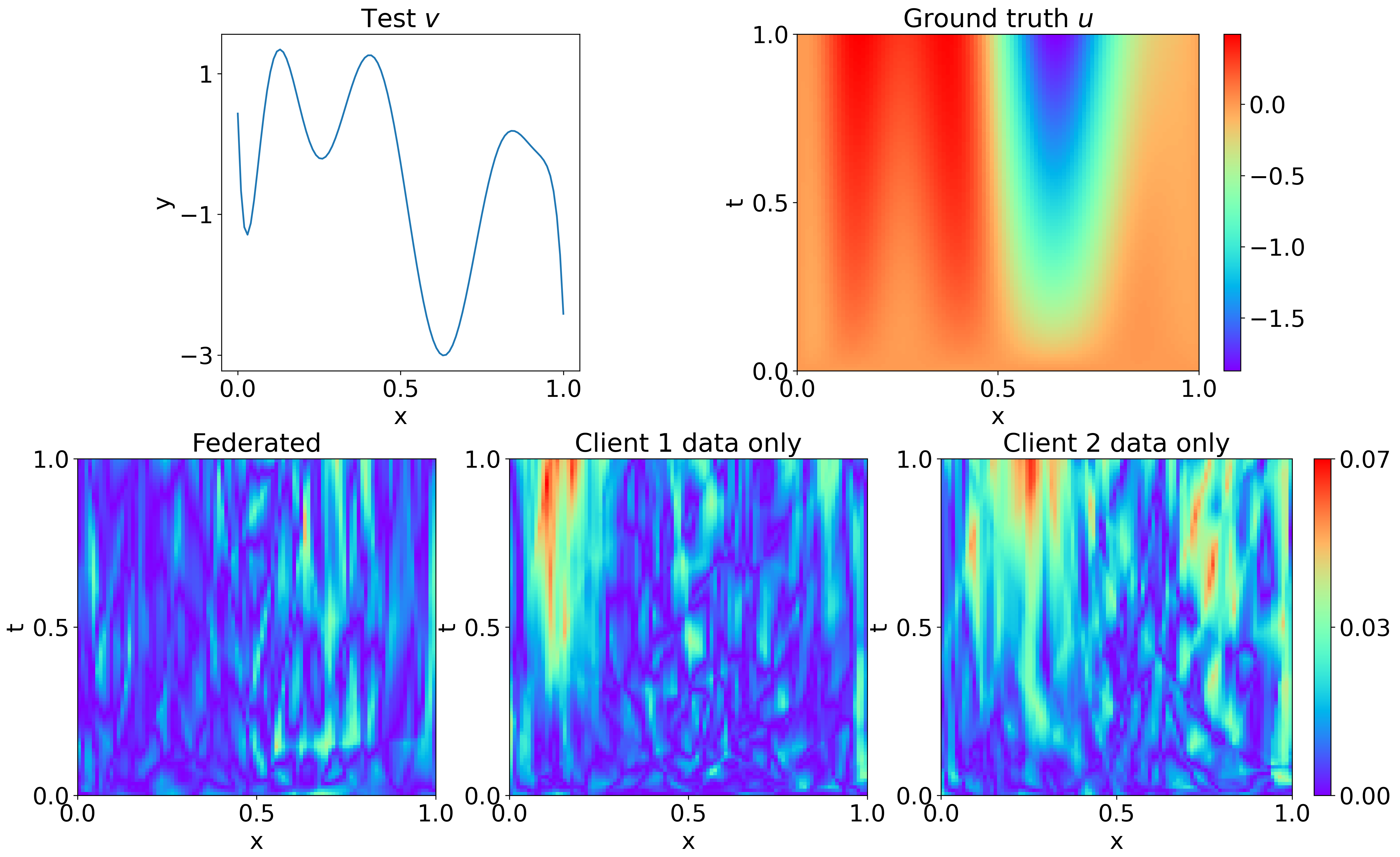}
    \caption{\textbf{Learning the diffusion-reaction equation in Section~\ref{subsec:dr}.} The top figures are examples of testing function $v(x)$ and ground truth solution $u(x,t)$. The bottom is an example of a federated model and two extrapolation baselines for $n = 10$. The first column is the absolute error for FedDeepONet, and the rest are the absolute error of extrapolation baselines.}
    \label{fig:deeponet_dr}
\end{figure}

We sampled the input functions $v(x)$ for testing from the Chebyshev functional space with $n=10$, and one example of $v(x)$ and corresponding solution $u(x,t)$ is visualized in the top row of Fig.~\ref{fig:deeponet_dr}. The bottom figures are the absolute errors of the predictions from FedDeepONet and extrapolation models, where the federated model outperforms the extrapolating baselines that are only trained locally. Besides, the error plot demonstrates that the performance of training only with the second client is worse than the training with the first client data, and an intuitive explanation can be the insufficient knowledge of the polynomial functional space by having nonzero terms in inverse order. This phenomenon indicates that the influence of local domain knowledge, i.e., how local data can represent the entire learning space, is more significant to the final performance of the federated model in the operator learning setting. 


\subsection{Burgers' equation}
\label{subsec:burgers}

Finally, we consider the Burgers' equation defined by
\begin{equation*}
    \frac{\partial u}{\partial t} + u \frac{\partial u}{\partial x} = \nu \frac{\partial^2 u}{\partial x^2}, \quad x \in [0,1], \quad t \in [0,1],
\end{equation*}
with a periodic boundary condition and an initial condition $u_0(x)=v(x)$. In this numerical experiment, $\nu$ is set at 0.1. The objective is to learn the operator mapping from initial condition $v(x)$ to the solution $u(x, t)$. The periodic function $v(x)$ is sampled from the Chebyshev polynomial functional space with $M=10$. Models are trained with 200 functions in total and tested on 500 functions. The architecture of DeepONet is the same as the diffusion-reaction equation.

\begin{figure}[htbp]
    \centering
    \includegraphics[width=\textwidth]{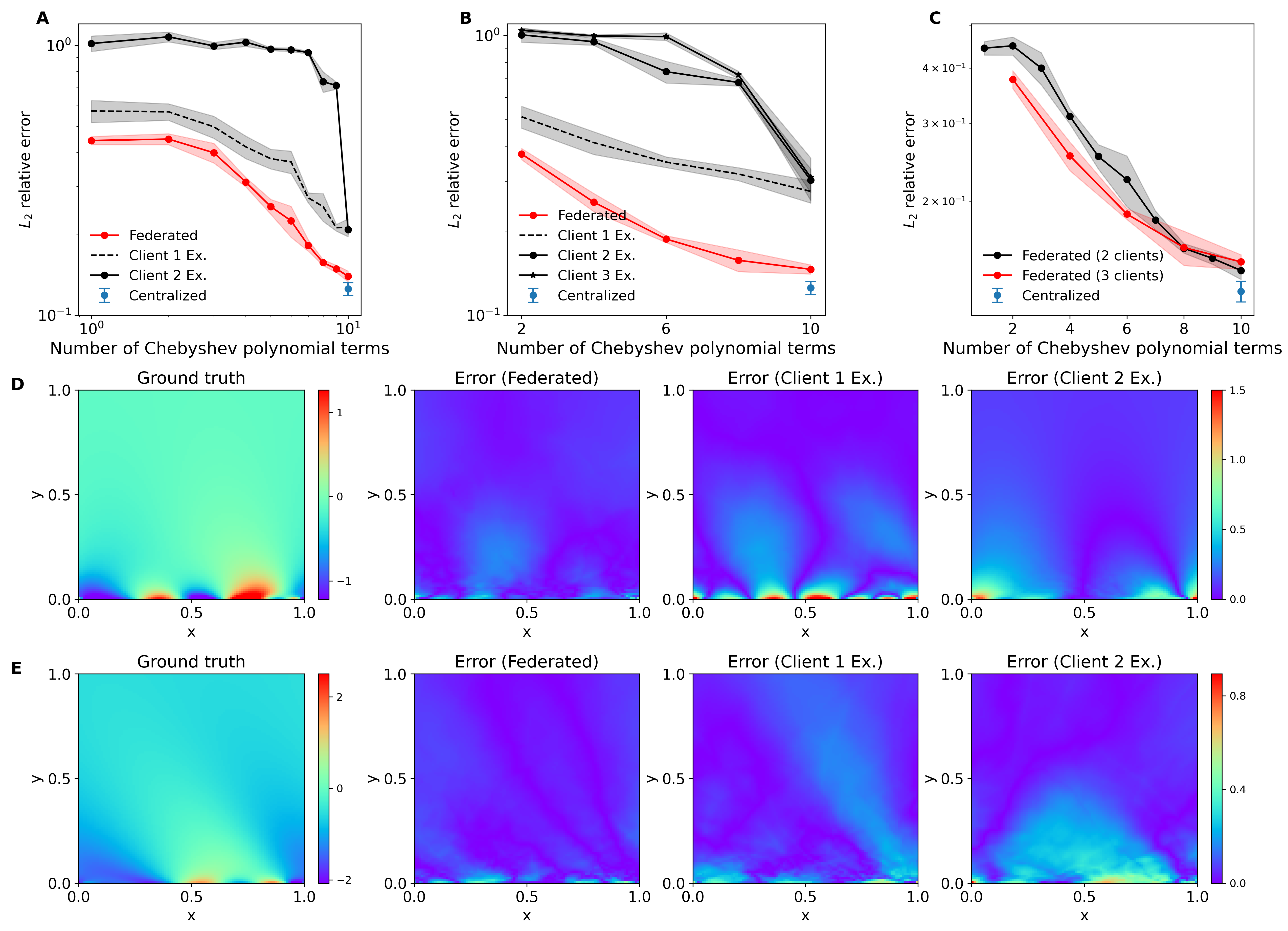}
    \caption{\textbf{Learning the solution operator of Burgers' equation in Section~\ref{subsec:burgers}.} (\textbf{A}) $L_2$ relative error for different numbers of nonzero Chebyshev polynomial terms for two clients. (\textbf{B}) $L_2$ relative error for three clients' different numbers of nonzero Chebyshev polynomial terms. (\textbf{C}) Comparison of
    federated model for 2 and 3 clients. (\textbf{D} and \textbf{E}) Examples of
    federated model and two extrapolation baselines for (D) $n=5$ and (E) $n=10$. The first column is the ground truth solution of one testing function, and the rest are absolute errors for FedDeepONet and extrapolation baselines.}
    \label{fig:deeponet_burgers}
\end{figure}

The iid level increases as the number of nonzero terms increases, and the $L_2$ relative error decreases (Fig.~\ref{fig:deeponet_burgers}A). The improvement in error has a sharp decrease for client 2's extrapolating scenarios but is relatively smooth for client 1, which is different from the diffusion reaction example. One reason is that for Burgers' equation with large viscosity, the diffusive effects become more pronounced, and the system's overall behavior becomes more diffusive, as shown in two examples of high and low data heterogeneity (Figs.~\ref{fig:deeponet_burgers}D and E). The dominant influence of viscosity promotes a fast decay in the solution and thus undermines the effect of different input functional spaces. 

Moreover, we consider a multi-client scenario for Burgers' equation, where the data generation and non-iid control follow Sec.~\ref{subsubsec:operator}. For multi-client cases, the error trend is consistent with previous experiments in that increasing the number of terms results in a more minor $L_2$ relative error (Fig.~\ref{fig:deeponet_burgers}B). Besides, the second client and third client exhibit similar error trends, both of which are worse than the first client. This demonstrates that the lack of knowledge greatly influences extrapolating tasks but can be mediated with federated learning. Fig.~\ref{fig:deeponet_burgers}C compares federated learning for two- and three-client settings. The performance of the federated model trained with three clients outperforms the federated model only with two clients for the same number of Chebyshev polynomial terms per client. The three-client setting has more information by introducing the middle nonzero terms. Moreover, as the number of terms increases, the performance of two and three clients converges to the centralized DeepONet.

\subsection{Insensitivity of accuracy to communication frequency}
\label{subsec: fed_deeponet_insensitivity}
To further explore the property and advantage of combining federated learning with operator learning, we tested the performance of FedDeepONet with respect to the different number of local iterations $E$ for learning the antiderivative operator and the diffusion-reaction operator. Given a fixed total number of iterations, we can control the communication frequency by tuning the local iterations, where this setting guarantees that for larger $E$, the required number of communication rounds is smaller. The hyperparameters for networks and optimizers are the same as in previous experiments. To explore the relationship between the accuracy of FedDeepONet and various communication frequencies, $E$ is set to be $ [1,2,5,10,20,50,100,200,500,1000]$ with the total number of iterations fixed to be $50000$. By doing so, local iterations can cover almost identical scenarios to centralized training ($E=1$) and huge cases ($E=1000$) and demonstrate the property of FedDeepONet more comprehensively. 

\begin{figure}[htbp]
    \centering
    \includegraphics[width=\textwidth]{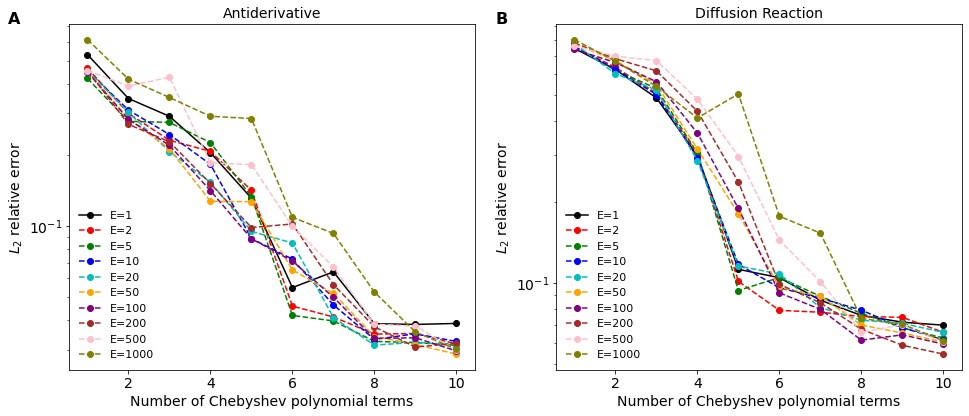}
    \caption{\textbf{Learning operators with respect to different numbers of local iterations. } $L_2$ relative error for different numbers of nonzero Chebyshev polynomial terms when training with different local epochs ($E$). (\textbf{A}) Antiderivative operator in Sec.~\ref{subsec:ad}. (\textbf{B}) Diffusion-reaction equation in Sec.~\ref{subsec:dr}.}
    \label{fig:insensitivity}
\end{figure}

The performances of different $E$ are very similar, even for $E$ as large as 1000, for both learning antiderivative operator and diffusion reaction operator (Figs.~\ref{fig:insensitivity}A and B). Empirical results validate that federated operator learning can achieve a similar level of accuracy regardless of the number of local updates and thus is shown insensitive to the communication frequency. This invariant property can be practically meaningful. In real-life applications, the communication round composed of one aggregation and one broadcasting usually counts for a significant portion of the computational cost. With the insensitivity of accuracy to communication frequency, the FedDeepONet can be efficiently trained with significant local iterations and a small number of communications to achieve satisfactory accuracy.

%% file: content/Conclusion.tex
\section{Conclusions}

This paper presents the first systematic study of federated scientific machine learning (FedSciML) to approximate functions and solve differential equations. We first propose different data generation methods corresponding to diverse problem settings, where synthetic data can be generated with different non-iidness to accommodate the decentralized training scenario. Subsequently, we provide a quantitative measure of the level of data heterogeneity by using 1-Wasserstein distance and validate that smaller $W_1$ distance, i.e., decreasing non-iid level, will result in smaller $L_2$ relative errors for federated models in all tasks. We also found that as federated models outperform the baseline models that are only trained on local datasets without communication with other clients, and as $W_1$ distance becomes small, such as $0.005$, the performance of federated models converge to the centralized baseline model for both forward and inverse problems. 

To better interpret the performance of the federated model compared with the centralized model, we introduce the generalized weight divergence to quantify the difference and analyze the upper bound of weight divergence for federated models used in this paper. In the later numerical experiment, we calculated the weight divergence in different layers for different non-iid levels. We demonstrated that weight divergence exhibits a consistent trend with error change in function approximation and FedPINN. However, this tendency is not evident for FedDeepONet, which may result from the different settings of operator learning and distinctive methods to control non-iid. Apart from the two-client scenario, we also study the performance of multiple clients and utilize the mean pairwise $W_1$ distance as the quantifier of data heterogeneity. The results show a similar trend as the two-client case despite the error becoming more extensive as the number of clients increases, which we conjecture is caused by the increasing divergence among clients.  

This study is the first attempt to understand and build the FedSciML framework, and more work should be done theoretically and computationally. This study considers the FedAvg method in aggregating local model parameters. Still, the effect of different weights on local models and even adaptive updating algorithms can be investigated in future work. Although we provide a theoretical study of the upper bound in this paper, more theoretical analysis of FedSciML is required in future work.
\label{sec:conclusion}

%% file: content/Appendix.tex
\section{Abbreviations and notations}
\label{si:notation}

Table~\ref{tab:notation} lists the main abbreviations and notations used throughout this paper.

\begin{table}[htbp]
    \centering
    \caption{\textbf{Abbreviations and notations.}}
    \vspace{2pt}
    \label{tab:notation}
    \begin{tabular}{ll}
        \toprule
        $\alpha_i$ & coefficient of $i$-th term in Chebyshev polynomials\\
		$\mathcal{G}$ & operator to learn\\
		$\Omega$ & domain of the PDE\\
            $D$ & dataset of the federated model\\
            $D_k$ & dataset of the $k$-th client of the federated model\\
            $\mathcal{E}_{WD}^{l, E}$ & weight divergence at the $l$-th global epoch and $E$-th local epoch\\
		$\{x_1, x_2, \cdots, x_m\}$ & scattered sensors\\
		$[v(x_1), v(x_2), \cdots, v(x_m)]$ & input of branch network\\
		$[b_1(v), b_2(v), \cdots, b_p(v)]$ & output of branch network\\
		$\xi$ & input of trunk network\\
		$[t_1(\xi), t_2(\xi),\cdots, t_p(\xi)]$ & outputs of trunk network, where $p$ is the number of neurons \\
		$\mathcal{F}$ & governing PDEs and/or physical constraints\\
            $\mathcal{L}$ & loss of the federated model\\
            $\mathcal{L}_k$ & loss of the $k$-th client of the federated model\\
            $\ell$ & point-wise squared error loss\\
        $K$ & number of clients\\
        $E$ & number of local epochs\\
        $l$ & number of global epochs\\
        $N$ & total number of samples \\
        $N_k$ & number of samples for client $k$\\
        \multirow{2}{*}{$n$} & number of subdomains in function approximation and PINNs\\
        & number of nonzero Chebyshev polynomial terms in operator learning\\
	\bottomrule
    \end{tabular}
\end{table}

\section{Neural network architectures}
\label{si:arch}

Table~\ref{tab:arch} provides the summary of network size, activation function, and training epochs in experiments in Sections~\ref{sec:func_approx}, \ref{sec:pinn} and \ref{sec:deeponet}. For the last three examples of training FedDeepONet, the size of hidden layers is the same for both trunk and branch nets, where the first layer's dimension depends on the input dimension and number of sensors.

\begin{table}[htbp]
    \centering
    \caption{\textbf{Network architectures and training settings.}}
    \vspace{2pt}
    \label{tab:arch}
    \begin{tabular}{l | cccccc}
        \toprule
        Examples & Width & Depth & Activation & Local epochs & Global epochs \\
        \hline
         Gramacy\&Lee (Sec.~\ref{subsec:gramacy})& 3 & 64 & tanh & 5 & 3000 \\
         Schaffer (Sec.~\ref{subsec:schaffer})& 3 & 64 & tanh & 5 & 3000  \\
         1D Poisson (Sec.~\ref{subsec:1dpoisson})& 3 & 20 & tanh & 5 & 1000 \\
         2D Helmholtz (Sec.~\ref{subsec:helmholtz})& 3 & 64 & sine & 5 & 2000  \\
         Allen-Cahn (Sec.~\ref{subsec:allencahn})& 3 & 64 & sine & 5 & 10000  \\
         Navier-Stokes (Inverse) (Sec.~\ref{subsec:ns})& 6 & 50 & tanh & 1 &  40000 \\
         Diffusion-reaction (Inverse) (Sec.~\ref{subsec:inverse_dr})& 3 & 20 & tanh & 5 & 20000  \\
         Antiderivative (Sec.~\ref{subsec:ad}) & 2 & 40 & ReLU & 5 & 10000  \\
         Diffusion-reaction (Sec.~\ref{subsec:dr})& 3 & 100 & ReLU & 5 & 10000   \\
         Burgers' (Sec.~\ref{subsec:burgers})& 2 & 64 & ReLU & 5 & 10000   \\
         \bottomrule
    \end{tabular}
    \label{tab:my_label}
\end{table}

\section{Proof to Theorem~\ref{thm:temp}}
\label{appendixD}
\begin{proof}
We use the notation $\theta^i$ to represent the weights of the centralized model at the $i$-th epoch, and $\theta_k^i$ represent the weights of the federated $k$-th client model at the $i$-th local epoch.

We have $\theta^0 = \theta_k^0, \forall k$ due to the same initialization of the centralized and federated models. Considering the one-epoch update of the centralized model, we have
\begin{align}
    \theta^{j+1} &= \theta^j - \eta \nabla \mathcal{L}(\theta^j; D) \nonumber\\
    &= \theta^j - \frac{\eta}{N}\sum_{d_i \in D} \nabla \ell(\theta^j; d_i). \label{equation:central_update}
\end{align}
Note Eq.~\eqref{equation:central_update} holds for any epoch $i$, we can sum up both sides of the equations for $j = 0, 1, \dots, E-1$ to get
\begin{equation*}
    \label{equation:central_to_0}
    \theta^E = \theta^0 - \frac{\eta}{N}\sum_{d_i \in D}\sum_{j=0}^{E-1} \nabla \ell(\theta^j; d_i).
\end{equation*}
Similarly, for the $k$-th client, the expression of the local model after $E$ local epoch is
\begin{equation*}
    \theta^E_k = \theta^0_k - \frac{\eta}{N_k}\sum_{d_i \in D_k}\sum_{j=0}^{E-1} \nabla \ell(\theta^j_k; d_i).
\end{equation*}
Hence, we can compute the norm of the difference in the weights between the centralized model and the client model by
\begin{align} \label{equation:normdiff}
    \left\Vert\theta^E - \theta^E_k\right\Vert &= \left\Vert \frac{\eta}{N}\sum_{d_i \in D}\sum_{j=0}^{E-1} \nabla \ell(\theta^j; d_i) - \frac{\eta}{N_k}\sum_{d_i \in D_k}\sum_{j=0}^{E-1} \nabla \ell(\theta^j_k; d_i)\right\Vert \nonumber \\
    &\leq \eta \sum_{j=0}^{E-1}\left\Vert\frac{1}{N}\sum_{d_i \in D}\nabla \ell(\theta^j; d_i) - \frac{1}{N_k}\sum_{d_i \in D_k}\nabla \ell(\theta^j_k; d_i)\right\Vert.
\end{align}
Apply the norm inequalities to the quantity $||\frac{1}{N}\sum_{d_i \in D}\nabla \ell(\theta^j; d_i)||$. Based on the assumption~\ref{assumption:bounded_grad}, we bound the gradients by a constant:
\begin{align}
    \left\Vert\frac{1}{N}\sum_{d_i \in D}\nabla \ell(\theta^j; d_i)\right\Vert &\leq \frac{1}{N}\sum_{d_i \in D}\left\Vert\nabla \ell(\theta^j; d_i)\right\Vert \nonumber\\
    &\leq \frac{1}{N}\sum_{d_i \in D}M\nonumber\\
    &= M.
    \label{eqn:bounded_sum_grad}
\end{align}
A similar conclusion holds for $||\frac{1}{N_k}\sum_{d_i \in D_k}\nabla \ell(\theta^j_k; d_i)||$:
\begin{equation}
    \left\Vert\frac{1}{N_k}\sum_{d_i \in D_k}\nabla \ell(\theta^j_k; d_i)\right\Vert\leq M \label{eqn:bounded_sum_grad_2}.
\end{equation}
Consequently, we can improve Eq.~\eqref{equation:normdiff} by Eq.~\eqref{eqn:bounded_sum_grad} and Eq.~\eqref{eqn:bounded_sum_grad_2}:
\begin{align*}
    \left\Vert\theta^E - \theta^E_k\right\Vert &\leq \eta \sum_{j=0}^{E-1}\left\Vert\frac{1}{N}\sum_{d_i \in D}\nabla \ell(\theta^j; d_i) - \frac{1}{N_k}\sum_{d_i \in D_k}\nabla \ell(\theta^j_k; d_i)\right\Vert \nonumber \\
    &\leq \eta \sum_{j=0}^{E-1}\left(\left\Vert\frac{1}{N}\sum_{d_i \in D}\nabla \ell(\theta^j; d_i)\right\Vert + \left\Vert\frac{1}{N_k}\sum_{d_i \in D_k}\nabla \ell(\theta^j_k; d_i)\right\Vert\right) \nonumber \\
    &\leq\eta \sum_{j=0}^{E-1}2M \nonumber \\
    &=2\eta EM.
\end{align*}

Now we look back to the weight divergence $\mathcal{E}_{\text{WD}}^{1,E}$. We have
\begin{align*}
    \mathcal{E}_{\text{WD}}^{1,E} &= \left\Vert \theta^E - \sum_{k=1}^K\frac{N_k}{N}\theta_k^{E}\right\Vert\nonumber\\
    &\leq \sum_{k=1}^K \frac{N_k}{N}\left\Vert\theta^E - \theta_k^E\right\Vert\nonumber \\
    &\leq 2\eta ME.
\end{align*}
The upper bound does not depend on the global epoch as long as the centralized and federated models share the same initialization. Hence, for $\mathcal{E}_{\text{WD}}^{l,E}$ with arbitrary global epoch $l$, it is easy to get
\begin{align*}
    \mathcal{E}_{\text{WD}}^{l,E} &\leq l\cdot \mathcal{E}_{\text{WD}}^{1,E}\nonumber\\
    &\leq 2\eta MEl.
\end{align*}

\end{proof}